\documentclass{article}

\usepackage[preprint]{neurips_2025}

\usepackage[utf8]{inputenc} 
\usepackage[T1]{fontenc}    
\usepackage[colorlinks,citecolor=gray]{hyperref}       
\usepackage{url}            
\usepackage{booktabs}       
\usepackage{amsfonts}       
\usepackage{nicefrac}       
\usepackage{microtype}      
\usepackage{xcolor}         

\usepackage{graphicx}
\usepackage{wrapfig}
\usepackage{caption}       
\usepackage{subcaption}
\usepackage{amsmath}
\usepackage{amssymb}
\usepackage{mathtools}
\usepackage{amsthm}
\usepackage{multirow}
\usepackage{bbm}
\usepackage{algorithm}
\usepackage{algorithmic}
\usepackage{setspace}
\usepackage{units}
\usepackage[capitalize,noabbrev]{cleveref}

\theoremstyle{plain}
\newtheorem{theorem}{Theorem}[section]

\title{Closing the Gap between TD Learning and Supervised Learning with $Q$-Conditioned Maximization}

\author{
Xing Lei$^{1}$ \quad \quad \And
Zifeng Zhuang$^{2}$\quad \quad \And
Shentao Yang$^{3}$\quad \quad \And
Sheng Xu$^{4}$\quad \quad \And
Yunhao Luo$^{5}$ \quad \quad \And
Fei Shen$^{6}$\quad \quad \And
Wenyan Yang$^{7}$\quad \quad \And
Xuetao Zhang$^{1,\dag}$ \quad \quad \And
Donglin Wang$^{2,\dag}$ \quad \quad \And\\
  $^{1}$Xi'an Jiaotong University\quad $^{2}$Westlake University\quad $^{3}$University of Texas at Austin\\
  $^{4}$The Chinese University of Hong Kong, Shenzhen\\
  $^{5}$Georgia Institute of Technology\quad
  $^{6}$National University of Singapore\\
  $^{7}$Aalto University School of Electrical Engineering, Aalto University\\
  \texttt{leixing@stu.xjtu.edu.cn}
}

\begin{document}

\maketitle
\renewcommand{\thefootnote}{\relax}
\footnotetext{\textsuperscript{$^\dag$}Corresponding Author.}

\begin{abstract}
Recently, supervised learning (SL) methodology has emerged as an effective approach for offline reinforcement learning (RL) due to their simplicity, stability, and efficiency. However, recent studies show that SL methods lack the trajectory stitching capability,  typically associated with temporal difference (TD)-based approaches. A question naturally surfaces: \textit{How can we endow SL methods with stitching capability and close its performance gap with TD learning?}
To answer this question, we introduce $Q$-conditioned maximization supervised learning for offline goal-conditioned RL, which enhances SL with the stitching capability through $Q$-conditioned policy and $Q$-conditioned maximization. Concretely, we propose \textbf{G}oal-\textbf{C}onditioned \textbf{\textit{Rein}}forced \textbf{S}upervised \textbf{L}earning (\textbf{GC\textit{Rein}SL}), which consists of (1) estimating the $Q$-function by Normalizing Flows from the offline dataset and (2) finding the maximum $Q$-value within the data support by integrating $Q$-function maximization with Expectile Regression. 
In inference time, our policy chooses optimal actions based on such a maximum $Q$-value.
Experimental results from stitching evaluations on offline RL datasets demonstrate that our method outperforms prior SL approaches with stitching capabilities and goal data augmentation techniques.
\end{abstract}

\section{Introduction}
Several recent papers reframes reinforcement learning (RL) as a pure supervised learning (SL) problem \citep{schmidhuber2020reinforcement,chen2021decision,emmons2021rvs,ghosh2021learning}, which has gained attention due to its simplicity, stability and scalability \citep{lee2022multi}. They typically assign labels to state-action pairs in the offline dataset based on the derived future outcomes (e.g., achieving a goal \citep{ghosh2021learning} or a return \citep{chen2021decision}); then maximize the likelihood of these actions by treating them as optimal for producing the labeled outcomes.
These approaches, termed as outcome-conditioned behavioral cloning (OCBC), have demonstrated excellent results in offline RL \citep{fu2020d4rl,emmons2021rvs}.
Nevertheless, recent studies \citep{yang2023swapped,ghugare2024closing} has identified these SL methods as the lack of stitching capability \citep{ziebart2008maximum}.
This is primarily because they do not maximize the $Q$-value \citep{kim2024adaptive}. In contrast, temporal difference (TD)-based RL methods (e.g., CQL \citep{kumar2020conservative}, IQL \citep{kostrikov2021offline}) possess stitching capability by learning and maximizing a $Q$-function, though they frequently encounter instability and optimization challenges \citep{van2018deep, kumar2019stabilizing} due to bootstrapping and projection into a parameterized policy space while maximizing the $Q$-value. 

To get the benefit of both world, in this paper, we focus on enhancing the stitching capability of SL-based method in offline RL while maintaining OCBC's stability. Inspired by recent max-return sequence modeling \citep{zhuang2024reinformer}, we propose a $Q$-conditioned maximization supervised learning framework.
We aim to incorporate $Q$-value as a conditioning factor in OCBC to acquire stitching capability, using the predicted maximum \textit{in-distribution} \citep{kostrikov2021offline} $Q$-value to determine the optimal action during inference, where the $Q$-value is supported by the offline dataset and is estimated via expectile regression \citep{aigner1976estimation, sobotka2012geoadditive}.

Algorithmically, we present \textbf{G}oal-\textbf{C}onditioned \textbf{Rein}forced \textbf{S}upervised \textbf{L}earning (\textbf{GC\textit{Rein}SL}), which implements $Q$-conditioned maximization supervised learning for OCBC methods, instantiated via DT \citep{chen2021decision} and RvS \citep{emmons2021rvs}. 
\textbf{GC\textit{Rein}SL} first estimates the $Q$-value from the offline dataset using Normalizing Flows \citep{ghugare2025normalizing}, and subsequently estimate the maximum $Q$-value together with the OCBC policy training.
This two-stage pipeline remove the need for the unstable bootstapping in standard TD-based method in learning the optimal $Q$-value.
\textbf{GC\textit{Rein}SL} not only learns the mapping between $Q$-value and action in the dataset, but also estimates the highest attainable \textit{in-distribution} $Q$-value during inference.

Despite its simplicity, the effectiveness of \textbf{GC\textit{Rein}SL} is empirically demonstrated on offline goal-conditioned RL datasets that require stitching \citet{ghugare2024closing}, outperforming prior OCBC methods and goal data augmentation methods \citep{yang2023swapped,ghugare2024closing}.
Furthermore, we extend our approach to return-conditioned RL without an explicit goal state, and compare it with state-of-the-art (SOTA) sequence modeling RL methods. Results on D4RL Antmaze \citep{fu2020d4rl} datasets show that our method continues to outperform related methods that also perform stitching. Theoretical and experimental evidence further indicates that our \textbf{GC\textit{Rein}SL} effectively closes the gap between OCBC and TD-based methods.
\section{Related Work}
\label{sec:prior-work}
The concept of trajectory stitching, as discussed by \citet{ziebart2008maximum}, is a characteristic property of TD-learning methods \citep{kumar2020conservative,kostrikov2021offline}, which employ dynamic programming. This property enables these methods to integrate data from diverse trajectories, thereby improving their ability to handle complex tasks by effectively utilizing available data \citep{cheikhi2023statistical}.
On the other hand, most SL-based methods, such as DT \citep{chen2021decision} and RvS \citep{emmons2021rvs}, lack this capability.
\citet{kumar2022offline,yang2023swapped} provide extensive experiments where SL algorithms do not perform stitching and 
\citet{ghugare2024closing} also demonstrates this from the perspective of combinatorial generalisation. Then they propose goal data augmentation for SL, yet these methods may struggle with correctly selecting augmented goal, such as unreachable goals \citep{yang2023swapped}. Unlike these methods, we first present an illustrative example to demonstrate that the SL approach lacks stitching capability. Subsequently, we enhance the stitching ability of SL by embedding the goal-reaching probability from the GCRL objective and maximizing it.

We further observe that several supervised learning methods \citep{jiang2023efficient, zeng2023goal, kim2024adaptive} demonstrate competitive performance in stitching tasks. However, unlike these approaches, our method \textbf{eliminates reliance on model-based mechanisms or dynamic programming for learning TD Q-value, instead leveraging the Normalizing Flows to estimate the Monte Carlo Q-value}. Conversely, other supervised learning methods, akin to our framework \citep{yamagata2023q,wu2023elastic,zhuang2024reinformer,wang2024critic}, have made like \textbf{one-step RL} \citep{brandfonbrener2021offline,zhuang2023behavior} in enabling OCBC to demonstrate stitching properties; nevertheless, their capability remains constrained. Our proposed \textbf{GC\textit{Rein}SL} effectively mitigates this limitation through maximize Monte Carlo Q-value.
\section{Preliminaries}\label{sc:prelimi}
\subsection{Goal-conditioned RL in Controlled Markov Process} \label{sc:2.1}
We study the problem of goal-conditioned RL in a controlled Markov process with states $s \in \mathcal{S}$, actions $a \in \mathcal{A}$. The dynamics is $p(s' \mid s, a)$, the initial state distribution is $p_0(s_0)$, the discount factor is $\gamma$,
and a reward function $r(s,a,g)$ for each goal. 
The goal-conditioned policy $\pi(a \mid s,g)$ is conditioned on a pair of state and goal $s,g\in\mathcal{S} \times \mathcal{G}$. 

For a policy $\pi$, we denote the $t$-step state distribution $p_t^\pi(s_t \mid s_0, a_0)$ as the distribution of
states $t$ steps in the future given the initial state $s_0$ and action $a_0$.
We can then define the discounted state occupancy distribution as: 
\begin{equation}\label{eq:dso_2}
p_+^\pi(s_{t+}\mid s,a) \triangleq (1-\gamma) \sum_{t=0}^{\infty} \gamma^{t} p_t^\pi(s_{t+} \mid s,a),
\end{equation}
where $s_{t+}$ is the dummy variable that specifies a future state corresponding to the discounted state occupancy distribution. 
For a given distribution over goals $g\sim p_{\mathcal{G}}(g)$, the objective of the policy $\pi$ is to maximize the probability of reaching the goal $g$ in the future:
\begin{equation}
\label{eq:goal-gradient}
    \max_{\pi(\cdot|\cdot,\cdot)}\mathbb{E}_{p_{0}(s_{0})p_{\mathcal{G}}(g)\pi(a_{0}|s_{0},g)}\big[p_+^{\pi}(g\mid s_{0},a_{0})\big].
\end{equation}
Following prior work \citep{eysenbach2020c, chane2021goal, blier2021learning, rudner2021outcome, eysenbach2022contrastive,bortkiewicz2025accelerating},
we define the goal-conditioned reward function $r(s,a,g)$ for each goal as the probability of reaching the goal at
the next time step:
\begin{equation}\label{reward}
r(s_t,a_t,g)\triangleq(1-\gamma)\gamma p(s_{t+1}=g\mid s_t,a_t).
\end{equation}
And the corresponding $Q$-function for a policy $\pi(\cdot\mid \cdot, g)$ can be defined as
\begin{equation} \label{eq:q}
Q^\pi(s,a,g)\triangleq\mathbb{E}_{\pi(\cdot\mid \cdot,g)}\left[\sum_{t=0}^{\infty}\gamma^{t}r(s_{t},a_{t},g)\mid\begin{smallmatrix}s_{0}=s,\\a_{0}=a\end{smallmatrix}\right].
\end{equation}
\textbf{Offline Setting.} Our work focuses on the offline RL setting
\citep{levine2020offline},
the agent can only access a static offline dataset $\mathcal{D}$
and cannot interact with the environment.
The offline dataset $\mathcal{D}$ can be collected from an unknown behavior policy $\beta$
\citep{levine2020offline,prudencio2023survey}.
We can express the offline dataset as ${\mathcal D}:=\{\tau_{i}\}_{i=1}^{N}$ \citep{ghugare2024closing},
where $\tau_{i} :=\left\{<s_{0}^{i},\eta_{0}^{i},a_{0}^{i},r_{0}^{i}>,<s_{1}^{i},\eta_1^{i},a_{1}^{i},r_{1}^{i}>,...,<s_{T}^{i},\eta_T^{i},a_{T}^{i},r_{T}^{i}>,g^i\right\}$ is the goal-conditioned trajectory and $N$ is the number of stored trajectories. In each $\tau_{i}$, 
$s_{0}^{i} \sim p_0(s_0)$,
and $\eta$ is the state’s corresponding  representation in the goal space calculated using $\eta_t = \phi(s_t^i)$, where $\phi:\mathcal{S}\rightarrow \mathcal{G}$
is a known state-to-goal mapping.
The desired goal $g^{i}$ is  randomly sampled from $p(g)$. 
It should be noted that trajectories may be unsuccessful (i.e,
$\eta^i_T \neq g^{i}$). Goal-conditioned methods often utilize $\eta_t, 0\leq t \leq T$ as relabeled goals $g$ for training.
\subsection{Outcome Conditional Behavioral Cloning (OCBC)} \label{sc:2.2}
We adopt a simple and popular class of goal-conditioned RL methods: outcome conditioned behavioral cloning~\citep{eysenbach2022imitating}, which encompasses DT~\citep{chen2021decision}, URL~\citep{schmidhuber2020reinforcement}, RvS~\citep{emmons2021rvs}, GCSL~\citep{ghosh2021learning} and so on. These SL methods take as input the offline dataset $\mathcal{D}$ and learn a goal-conditioned policy $\pi(a \mid s, g)$ using a maximum likelihood objective:
\begin{equation}
    \label{eq:objective}
    \max_{\pi(\cdot \mid \cdot, \cdot)} \mathbb{E}_{(s, a, g) \sim \mathcal{D}} \left[ \log \pi(a \mid s, g) \right].
\end{equation}
\section{Stitching in OCBC: Goal-reaching Probability-conditioned Maximization}\label{sc:4}
In the offline RL literature, trajectory stitching has garnered significant attention. Recent research by \citet{ghugare2024closing} interprets stitching from the perspective of combinatorial generalisation and demonstrates that OCBC methods lack the effective stitching capabilities. This finding is also corroborated experimentally by \citet{yang2023swapped}. Then they propose goal data augmentation methods that enhances the stitching performance of OCBC. Motivated by these prior works, we illustrate the limitations of OCBC in achieving stitching capability; however, unlike previous studies, we enable OCBC to acquire this capability by conditioning on the maximized goal-reaching probability.

\begin{figure}[h]
    \centering
    \vspace{-6pt}
    \begin{minipage}{0.325\linewidth}
        \centering
        \vspace{3pt}
		\centerline{\includegraphics[width=0.8\textwidth]{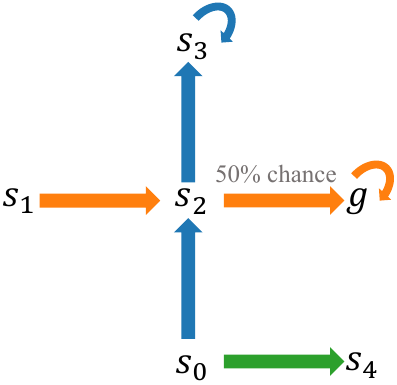}}
        \centerline{(a) Example MDP.}
	\end{minipage}
    \begin{minipage}{0.329\linewidth}
        \centering
        \vspace{3pt}
		\centerline{\includegraphics[width=\textwidth]{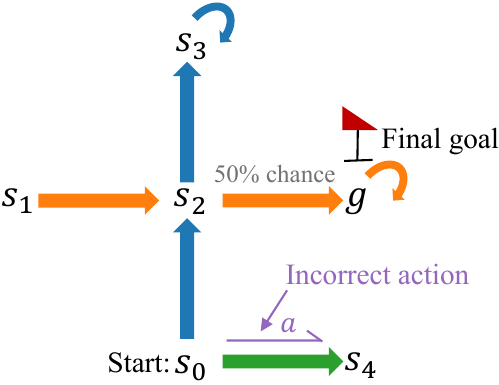}}
        \centerline{(b) OCBC fail to stitch.}
	\end{minipage}
     \begin{minipage}{0.329\linewidth}
        \centering
        \vspace{3pt}
		\centerline{\includegraphics[width=\textwidth]{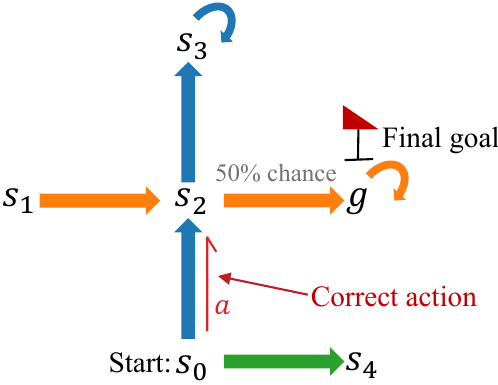}}
        \centerline{(c) Stitch.}
	\end{minipage}
    \vspace{-6pt}
    \caption{An illustrative example for stitching analysis. \textit{(a)} \textbf{Example MDP:} The MDP has five states, one goal and two actions (right $\textcolor{violet}{a\rightarrow}$ and up $\textcolor{red}{a\uparrow}$). One example offline dataset $\mathcal{D_{MDP}}$ contains two trajectories $\tau_1 = \{s_0, s_2, s_3\}$ and $\tau_2 = \{s_1, s_2, g\}$, distinguished by \textcolor{blue}{blue} and \textcolor{orange}{orange}. Another \textcolor{green}{green} trajectory $\tau_3 = \{s_0, s_4\}$ is not in $\mathcal{D_{MDP}}$.  \textit{(b)} \textbf{OCBC fails to stitch}: Given the start state $s_0$ and the final goal $g$, the classical OCBC policy tends to take the incorrect action (right, $\textcolor{violet}{a \rightarrow}$) that leads to undesired goal state $s_4$. \textit{(c)} \textbf{GC\textit{Rein}SL succeeds to stitch}: In contrast, given the $s_0$ and $g$, the \textbf{GC\textit{Rein}SL} policy is able to take the correct action (up, $\textcolor{red}{a \uparrow}$), causing $s_0 \rightarrow s_2 \rightarrow g.$}
    \vspace{-6pt}
    \label{maze}
\end{figure}

To demonstrate the lack of trajectory stitching in OCBC methods, consider the example in \cref{maze}: $s_0$ is the starting state, $g$ is the final goal. The example offline data $\mathcal{D}$ contains two trajectories $\tau_1 = \{s_0, s_2, s_3\}$ and $\tau_2 = \{s_1, s_2, g\}$. 
During inference, we expect that the policy can achieve the final goal $g$ given the start $s_0$. 
However, no trajectory in $\mathcal{D}$ goes directly from start $s_0$ to final goal $g$. 
In this case, starting from start $s_0$ and conditioned on $g$, the SL-based OCBC policy tends to take the wrong right action $\textcolor{violet}{a\rightarrow}$ because the policy believes the up action $\textcolor{red}{a \uparrow}$ will achieve the state $s_3$ rather than $g$ due the existence of the \textcolor{blue}{blue} trajectory $\tau_1$.

Ideally, the policy should stitch the existing trajectories and take one \textbf{stitched} trajectory $\tau^* = \left\{s_0, s_2, g\right\}$ to achieve $g$ from start $s_0$.
Dynamic programming based methods can propagate rewards through the backwards \textbf{stitch} path of $g \rightarrow s_2 \rightarrow s_0$ to output the correct action. 
Therefore, \citet{yang2023swapped,ghugare2024closing} propose an additional sampling of trajectory $\{s_0, g\}$ during the OCBC training phase and describe their approach as goal data augmentation.
In contrast, we additionally introduce a probability-conditioned policy, namely $\pi\left(a|s,g,P\right)$.
And during the inference phase, one proper probability $P^*$ is adopted to make this policy take the correct action.

To select a proper $P^*$, first, we denote $P(s,a,g)$ as the probability of reaching goal $g$ in the future by taking action $a$ from state $s$ (consistent with the probability definition in \cref{eq:goal-gradient}), it is evident that the following holds: $P(s_0,\textcolor{red}{a\uparrow},g)=\nicefrac{1}{2}$, 
and $P(s_0,\textcolor{violet}{a\rightarrow},g)=0$. 
When policy aims to achieve the final goal $g$ given the start $s_0$, we can use extra $P$ condition to guide the policy. 
Concretely, given the maximized conditional  $P^* = \max\left[P(s_0,\textcolor{red}{a\uparrow},g), P(s_0,\textcolor{violet}{a\rightarrow},g)\right] = \nicefrac{1}{2}$, the P-conditional policy $\pi\left(\cdot|s_0,g,P^*\right)$ will take the up action $\uparrow$ to achieve the desired goal-reaching probability.
\section{\textbf{GC\textit{Rein}SL}: \textbf{G}oal-\textbf{C}onditioned \textbf{\textit{Rein}}forced Supervised Learning}\label{sc:method}
From the perspective outlined in \cref{sc:4}, we aim to equip OCBC methods with the ability to maximize the expected probability of reaching the goal, as described in \cref{eq:goal-gradient}.
Recalling that the goal-reaching probability is equivalent to $Q$-function in GCRL (\cref{sec:5.1}), in \cref{sec:5.2}, we introduce the framework of $Q$-conditioned maximization supervised learning and theoretically demonstrate that this paradigm can achieve maximum $Q$-value without encountering the out-of-distribution (OOD) issue.
In \cref{sec:5.3}, we outline the practice implementation of our \textbf{GC\textit{Rein}SL}.
\subsection{The Relationship Between Goal-reaching Probability and $Q$-function} \label{sec:5.1}
\begin{theorem} [Rephrased from Proposition 1 of \citet{eysenbach2022contrastive} : probabilities $\rightarrow$ rewards]
\label{theorem:1}
The probability of reaching goal $g$ under the discounted state occupancy measure in \cref{eq:dso_2} is equivalent to the $Q$-function for the goal-conditioned reward function in \cref{eq:q}:
\begin{equation} \label{eq:q_prob}
p_+^\pi(s_{t+}=g \mid s,a) = Q^\pi(s,a,g).
\end{equation}
\end{theorem}
This theorem indicates that under the definition of reward in \cref{reward}, the goal-reaching probability $p_+^\pi(s_{t+}=g \mid s,a)$ is equivalent to a $Q$-function $Q^\pi(s,a,g)$. 

Translating probability into reward simplifies the analysis of goal-conditioned reinforcement learning (RL) and enables the use of probabilistic models, such as Conditional Variational Autoencoders (CVAE) \citep{sohn2015learning}, C-Learning \citep{eysenbach2020c}, Contrastive RL (CRL) \citep{eysenbach2022contrastive}, and Normalizing Flows \citep{ghugare2025normalizing}, for $Q$-function estimation. Given that Normalizing Flows can precisely compute this probability while reducing computational cost and complexity \citep{ghugare2025normalizing}, in \cref{sec:4.2}, we provide a detailed implementation for estimating the goal-reaching probability and $Q$-function using Normalizing Flows. This makes them particularly suitable for integration into our supervised learning framework, where accurate and efficient probability estimation is paramount for effective stitching. In \cref{sc:ablation study}, we discuss the impact of different estimators on the final performance.
\subsection{$Q$-conditioned maximization supervised learning} \label{sec:5.2}
Assume that we can accurately estimate the $Q$-function $Q^\beta(s,a,g)$ of the behavior policy $\beta$ for each state-action pair in the offline dataset (i.e., accurately obtain the goal-reaching probability for a given goal along the same trajectory), 
we aim to equip supervised learning with an additional objective of maximizing $Q$-function so as to obtain the maximum \textit{in-distribution} $Q$-value.
Then, during inference, the policy can select (near-) optimal action conditioned on the \textit{in-distribution} maximized $Q$-value.
Expectile regression \citep{newey1987asymmetric} is suitable to capture the upper distribution bound \citep{kostrikov2021offline,wu2023elastic,zhuang2024reinformer}, so we employ it as $Q$-function loss for estimating the maximum \textit{in-distribution} $Q$-value.

Specifically, the $Q$-function loss based on the expectile regression is as follows:
\begin{equation}\label{eq:expectile regression}
    \mathcal{L}^{m}_{\hat{Q}}=\mathbb{E}_{(s,a,g)\sim \mathcal{D}}\left[\left|m-\mathbbm{1} \left( \Delta Q < 0\right) \right|\Delta Q^2\right],
\end{equation}
here $\Delta Q = Q^\beta - \hat{Q}$ 
and $\hat{Q}$ is the predicted $Q$-value for the learned policy $\pi$ that can come from the supervised learning model 
(e.g., DT model can independently predict both the $Q$-value and the corresponding actions).
Here $m \in \left(0,1\right)$ is the hyperparameter of expectile regression.
When $m=0.5$, expectile regression reduces to the standard Mean Squared Error (MSE) loss.
However, when $m>0.5$, this asymmetric loss function places greater weight on $Q$-values larger than $\hat{Q}$.
In other words, the predicted $Q$-value $\hat{Q}(s,a)$ will approach larger $Q^{\beta}(s,a)$.

To reveal what the $Q$-function loss in \cref{eq:expectile regression} will learn and provide a formal explanation of its role, we present the following theorem:
\begin{theorem}\label{theorem:2}
Define $\mathbf{SG} \dot= \left(s,g,a,Q^{\beta}\right)$.
For $m\in\left(0,1\right)$, denote $\mathbf{Q}^m\left(\mathbf{SG}\right) = \arg \min_{\hat{Q}} \mathcal{L}_{\hat{Q}}^m\left(\mathbf{SG}\right)$,  we have
\begin{align*}
    \lim_{m\rightarrow 1} \mathbf{Q}^m\left(\mathbf{SG}\right) = Q_{\text{max}}\,, \: \forall s, g\,,
\end{align*}
where $Q_{\text{max}} = \max_{\mathbf{a} \sim \mathcal{D}} Q^{\beta}\left(s,a,g\right)$ denotes the maximum $Q$-value over all actions under $s$ in the offline dataset.
\end{theorem}
The proof is in \cref{pf:2}. It is crucial to note that $Q_{\text{max}}$ here refers to the maximum action value in the dataset, \textit{not the global maximum}, as the offline dataset may not contain the global maximum. \cref{theorem:2} indicates the loss $\mathcal{L}_{\hat Q}^m$ will make $\hat Q$ predict the maximum $Q$-value when $m\rightarrow 1$, which is similar to the objective of maximizing the $Q$-function in traditional RL.
\subsection{Practical Implementation}\label{sec:5.3}
\begin{figure*}[h]
\centering
\vspace{-6pt}
\centerline{\includegraphics[width=\textwidth]{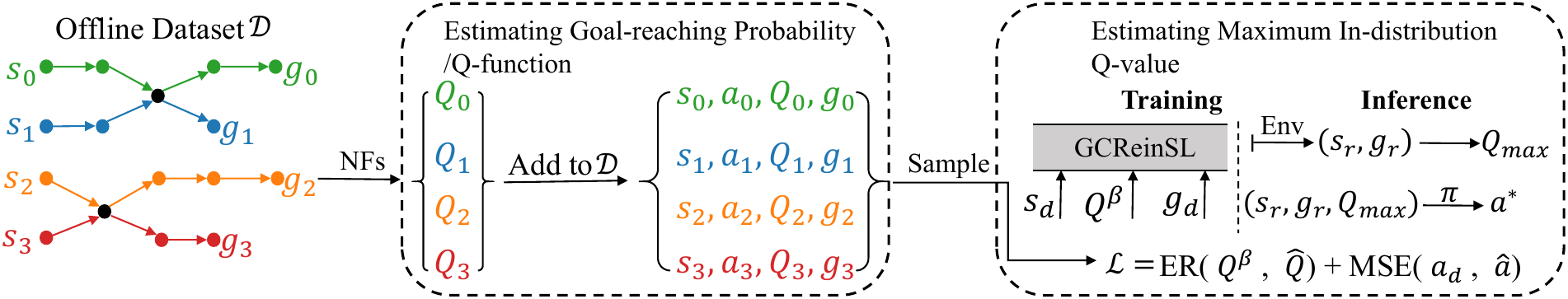}}
\caption{The overview of \textbf{GC\textit{Rein}SL} structure. $[s_0,s_1,s_2,s_3]\in s_d$, $[a_0,a_1,a_2,a_3]\in a_d$, $[g_0,g_1,g_2,g_3]\in g_d$ and $[Q_0,Q_1,Q_2,Q_3]\in Q^\beta$ come from offline data $\mathcal{D}$. $(s_r,g_r)$ come from environment. ER denotes Expectile Regression. $Q_{max}$ denotes \textit{in-distribution} max $Q$-value. $\hat{Q}$ and $\hat{a}$ represent the predicted $Q$-value and the output action of the model, respectively.
\textbf{Left:} The original offline dataset $\mathcal{D}$.
\textbf{Middle:} Normalizing Flows (NFs) as an estimator for the goal-reaching probability/$Q$-function.
\textbf{Right:} The \textbf{GC\textit{Rein}SL} model trains using the modified loss $\mathcal{L}$ and estimates the maximum $Q$-value during the inference phase to output the optimal action. Note that our policy here is a $Q$-conditioned policy $\pi(a|s,g,Q)$, which aligns with the definition provided in \cref{sc:4}.}
\vspace{-6pt}
\label{gcreinsl_overview}
\end{figure*}

Now, we will focus on the concrete implementation of
\textbf{GC\textit{Rein}SL}, including the component of goal-reaching probability/$Q$-function estimation and the requirement of estimating the maximum $Q$-value. The overall structure of \textbf{GC\textit{Rein}SL} is depicted in \cref{gcreinsl_overview}.

\subsubsection{Estimating goal-reaching probability/$Q$-function} \label{sec:4.2}
\begin{figure*}[h]
  \centering
  \vspace{-6pt}
  \includegraphics[width=0.8\linewidth]{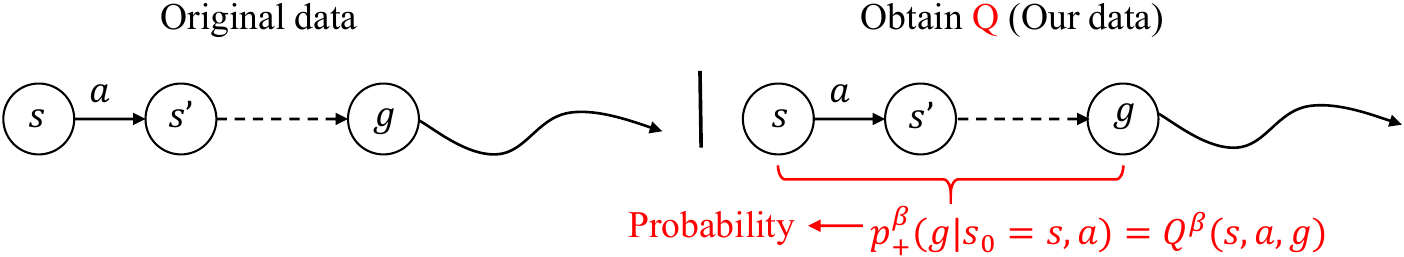}
  \caption{Estimating the $Q$-function of the behavior policy via Normalizing Flows. \textbf{Left:} Original offline trajectory, where the goal $g$ is reachable from the state $s$. \textbf{Right:} Normalizing Flows are trained to directly estimate the log-likelihood, $\log p_+^\beta(g \mid s_0=s,a)$. Note that $p_+^\beta(g \mid s_0=s,a)$ is exactly the goal-reaching probability for the behavior policy $\beta$.}
  \vspace{-6pt}
  \label{NF}
\end{figure*}

The central aim of goal-conditioned RL is to identify the best action for a given state and goal to maximize the chance of reaching the given goal. To achieve this, the first requirement of our method necessitates a precise estimation of the $Q$-function $Q^{\beta}(s,a,g)$ under the goals appeared in the dataset. Drawing on previous research \citep{zhai2024normalizing,ghugare2025normalizing} and \cref{theorem:1}, we employ Normalizing Flows to directly estimate the goal-reaching probability/$Q$-function. \cref{NF} succinctly illustrates this process.

We employ a conditional Normalizing Flow model $f_\psi : \mathcal{G} \times \mathcal{S} \times \mathcal{A} \rightarrow \mathcal{Z}$, which is an invertible neural network that maps the goal $g$ (conditioned on $s$ and $a$) to a latent variable $z$ in a base distribution (typically a standard Gaussian $\mathcal{N}(0, I)$). The probability density $p_\psi(g | s, a)$ is then given by the change of variables formula \citep{papamakarios2021normalizing}:
\begin{equation}
p_\psi(g|s,a)=p_{\mathcal{Z}}(f_\psi(g;s,a))\cdot\left|\det\frac{\partial f_\psi(g;s,a)}{\partial g}\right|,
\end{equation}
where $p_\mathcal{Z}$ is the density of the base distribution and the Jacobian determinant $\det \frac{\partial f_\psi}{\partial g}$ accounts for the volume change under the transformation.
Our architecture for $f_\psi$ builds upon the highly expressive yet efficient design proposed by \citet{ghugare2025normalizing}, which combines coupling layers \citep{dinh2017density} and linear flows \citep{kingma2018glow}. This design ensures that both the forward mapping $f_\psi$ and its inverse are computationally tractable, and the Jacobian determinant can be calculated efficiently.

We train the flow model $f_\psi$ via maximum likelihood estimation (MLE) on the offline dataset $\mathcal{D}$, maximizing the probability of observed future goals given their corresponding state-action pairs:
\begin{equation}\label{eqn:nfs}
\max_\psi\mathbb{E}_{(s,a,g)\sim\mathcal{D}}\left[\log p_\psi(g|s,a)\right].
\end{equation}
Once trained, the estimated $Q$-value for any $(s, a, g)$ tuple is obtained by evaluating the log-likelihood of the goal under our model:
\begin{equation}\label{eqn:nfq}
Q^\beta(s,a,g)=p_\theta(g|s,a).
\end{equation}
In \cref{ap:evaluate nf}, we discuss the accuracy of the Normalizing Flows in estimating this $Q^{\beta}(s,a,g)$.
\subsubsection{Estimating the maximum $Q$-value}
After estimating the $Q$-value using Normalizing Flows, we apply our \textbf{GC\textit{Rein}SL} loss for the OCBC to estimate the maximum values within the dataset. The $Q^{\beta}(s,a,g)$ values serve as additional conditioning factors in our policy during the training phase. Meanwhile, the  estimated maximum $Q$-value is used as an additional conditioning factor during inference.
\textbf{Training (\text{\textcolor{blue}{Integrating the expectile regression into the OCBC loss}}).}
Since our overall agent predicts both $Q$-value $\hat{Q}$ and action $\hat{a}$, its training loss consists of a $Q$-function loss (
\cref{eq:expectile regression}) and an action loss. For the action loss, we adopt the MSE loss function in OCBC. We use the same weight for these two loss function terms and therefore the total loss is:
\begin{equation}\label{eq:total_loss}
\mathcal{L}^{\textbf{GC\textit{Rein}SL}}_{\pi, \hat Q}=\mathbb{E}_{(s,a,g,Q^{\beta})\sim \mathcal{D}}\left[ \underbrace{\left\|a-\pi(s,g,\hat{Q})\right\|_{2}^{2}}_{\mathrm{OCBC}} +\underbrace{\left|m-\mathbbm{1} \left( \Delta Q < 0\right) \right|\Delta Q^2}_{\mathrm{\textit{in-distribution} ~max~Q-value}}\right],
\end{equation}
where $\Delta Q = Q^{\beta} - \hat{Q}$ and $m>0.5$ represents the hyperparameter of expectile regression.

\textbf{Inference (\text{\textcolor{blue}{Stitch}}).}
In classical $Q$-learning \citep{mnih2015human}, the optimal value function $Q^*$ can derive the optimal action $a^*$ given the current state. 
In the context of OCBC, we are therefore motivated to believe that the maximum $Q$-value can help the policy select the (near-)optimal actions. 
Note that the maximum $Q$-value in the offline dataset depends only on the state and goal, as action is ``reduced'' by the $\max$ operation.
The inference pipeline of the \textbf{GC\textit{Rein}SL} is summarized as follows:
\begin{align}
    \overset{\text{\textcolor{blue}{Env}}}{\longmapsto} \left(s_0,g_0\right) \xrightarrow{} \hat{Q}_0 \xrightarrow{\pi} a_0 \xrightarrow{\text{\textcolor{blue}{Env}}} \left(s_1,g_1\right) \xrightarrow{} \hat{Q}_1 \xrightarrow{\pi} a_1 \rightarrow \cdots
\end{align}
Specially, the environment initializes the state-goal pair $\left(s_0,g_0\right)$ and then our model predicts the maximum $Q$-value $\hat{Q}_0$ given current state-goal pair $\left(s_0,g_0\right)$.
Based on $\hat{Q}_0$ and $\left(s_0,g_0\right)$, $\pi_{\theta}$ selects an action $a_0$.
It is important to note that during inference time, the  pair of initial state and goal from the environment may corresponding to the inital state and goal of different trajectories in the offline dataset (like $\{s_0, g\}$ in \cref{sc:4}). In this case, our model can still output good actions by stitching together sub-trajectories from multiple trajectories in the dataset.
With $a_0$, the environment transitions to the next state $s_1$ and receive the new goal $g_1$.
In \cref{app:alg}, we present the model and algorithm details using DT \citep{chen2021decision} and RvS \citep{emmons2021rvs}  as the SL backbone.
\subsection{Comparison and Analysis}
\begin{wrapfigure}[17]{r}{0.45\linewidth}
    \vspace{-6pt}
    \centering
	\centerline{\includegraphics[width=0.45\textwidth]{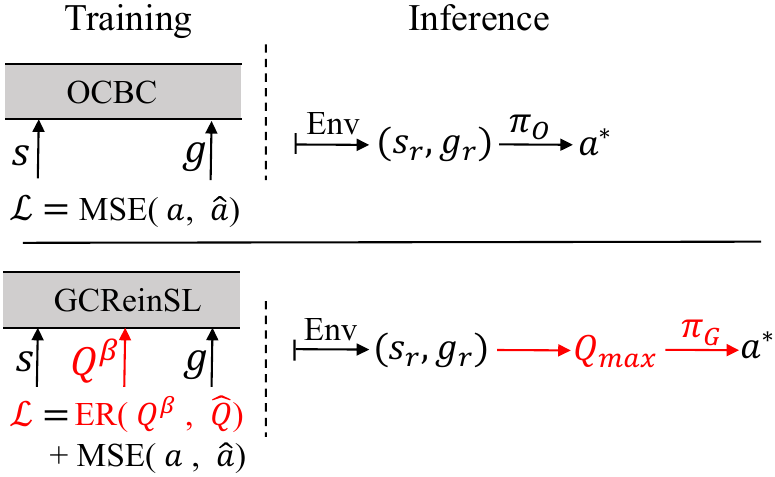}}
    \caption{
    \textbf{Left and Right at the Top:} OCBC. \textbf{Left and Right at the Bottom:} \textbf{GC\textit{Rein}SL}. $s$, $g$ and $Q^\beta$ are come from offline data $\mathcal{D}$. 
    $s_r$ and $g_r$ are come from environment. ER denotes Expectile Regression. The red section highlights the differences. 
    }
    \vspace{-6pt}
    \label{fig:comparasion results}
\end{wrapfigure}
To further clarify the differences between OCBC and our \textbf{GC\textit{Rein}SL}, as well as the benefits of our changes, we provide a comparison of OCBC and \textbf{GC\textit{Rein}SL} in \cref{fig:comparasion results}. We can observe that our \textbf{GC\textit{Rein}SL} introduces an additional conditioning factor, $Q^\beta$, and employs expectile regression loss to obtained the maximized \textit{in-distribution} $Q$-value ${Q}_{max}$. The inference process determines the optimal action $a^{*}$ by considering both the given state-goal pair $(s_r,g_r)$ and the model predicted $Q_{max}$. Note that the learned policy change from $\pi_{O}=\pi(a|s,g)$ in OCBC to $\pi_{G}=\pi(a|s,g, \textcolor{red}{Q})$ in \textbf{GC\textit{Rein}SL}.

Thanks to the additional conditioning on $Q^\beta$ and the maximization of $Q_{max}$, we can incorporate information from other trajectories to facilitate the stitching process. As shown in \cref{maze}, the agent can select the optimal action by maximizing the $Q$-value. Furthermore, our method does not require the unstable bootstrapping in learning the maximum $Q$-value, unlike TD-based method such as IQL \citep{kostrikov2021offline}, which needs to first learn a $Q^\beta$ before learning $\hat Q$. 
As an extra benefit over the TD-based method, the SL nature of our method removes the need for additional mechanisms to project the $Q$-maximizing policy to a parameterized policy space from which one can easily sample, such as CQL \citep{kumar2020conservative}, TD3+BC \citep{fujimoto2021minimalist} and IQL \citep{kostrikov2021offline}. In \cref{appendix: return-conditioned rl}, we discuss the extension  of \textbf{GC\textit{Rein}SL} to return-conditioned RL, where there is no concrete goal state.
\section{Experiments} \label{sec:experiments}
This section aims to address three key questions: 1) How do the stitching capabilities and degree of stitching of \textbf{GC\textit{Rein}SL} perform across different benchmarks? 2) How does \textbf{GC\textit{Rein}SL} behave under high-dimensional inputs? 3) When extended to return-conditioned RL, how does it compare to prior sequence modeling methods, and does it narrow the performance gap with TD-based methods?
\subsection{Experimental Setup}
To evaluate the stitching capability of \textbf{GC\textit{Rein}SL}, we employ the offline \citet{ghugare2024closing} datasets for goal-conditioned RL and D4RL \citep{fu2020d4rl} \texttt{Antmaze-v2} datasets for return-conditioned RL. We select RvS \citep{emmons2021rvs} and DT \citep{chen2021decision}, two competitive methods in OCBC, as baseline models for comparison. We compare \textbf{GC\textit{Rein}SL} with three categories of existing methods: 
(1) for goal data augmentation methods,
we include Swapped Goal Data Augmentation (SGDA) \citep{yang2023swapped} and Temporal Goal Data Augmentation (TGDA) \citep{ghugare2024closing} with DT and RvS;
for sequence modeling methods, we include Elastic Decision Transformer (EDT) \citep{wu2023elastic}, Critic-Guided Decision Transformer (CGDT) \citep{wang2024critic}, Max-Return Sequence Modeling (Reinformer) \citep{zhuang2024reinformer} and Q-value Regularized Transformer \citep{hu2024q} [QT (1-step)]; for TD-based RL methods, we include CQL \citep{kumar2020conservative} and IQL \citep{kostrikov2021offline}. See \cref{sc:baseline details} for more details of baselines.  
All experiments are conducted using five random seeds. Following the related original paper \citep{ghugare2024closing,zhuang2024reinformer}, we report the final mean success rate in goal-conditioned RL and the best score in return-conditioned RL experiments. Detailed implementations and hyperparameters are provided in  \cref{ap:exp details} and \cref{ap:hy}, respectively.
\subsection{Testing the Stitching Capability of \textbf{GC\textit{Rein}SL} in \texttt{Pointmaze} Datasets}

\begin{figure*}[t]
    \vspace{-6pt}
    \centering
	\begin{minipage}{\linewidth}
		\centerline{\includegraphics[width=8.5cm]{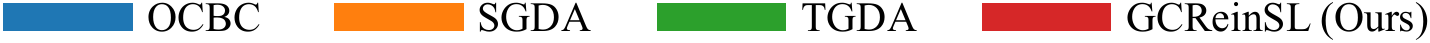}}
    \end{minipage}
    \begin{minipage}{0.32\linewidth}
		\centerline{\includegraphics[width=\textwidth]{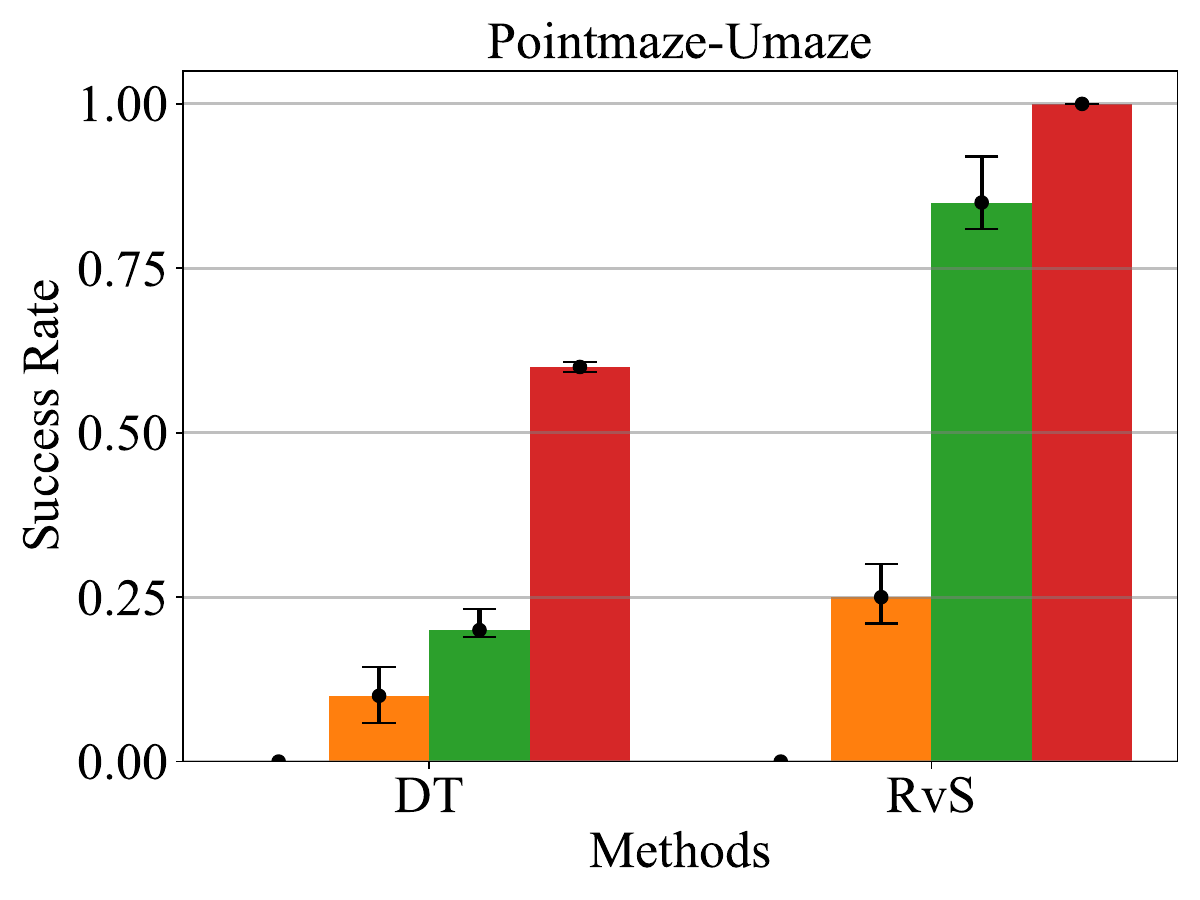}}
	\end{minipage}
    \begin{minipage}{0.32\linewidth}
		\centerline{\includegraphics[width=0.99\textwidth]{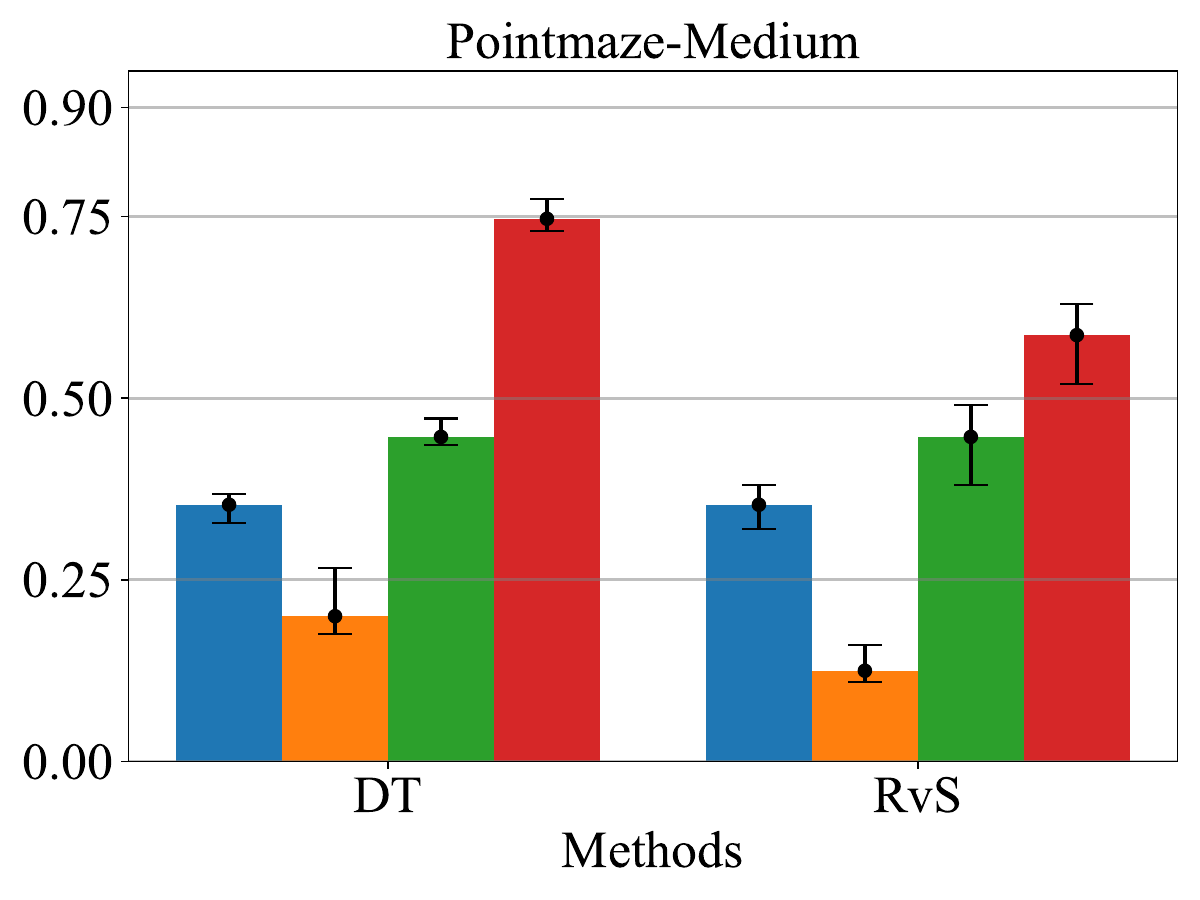}}
	\end{minipage}
	\begin{minipage}{0.32\linewidth}
		\centerline{\includegraphics[width=0.99\textwidth]{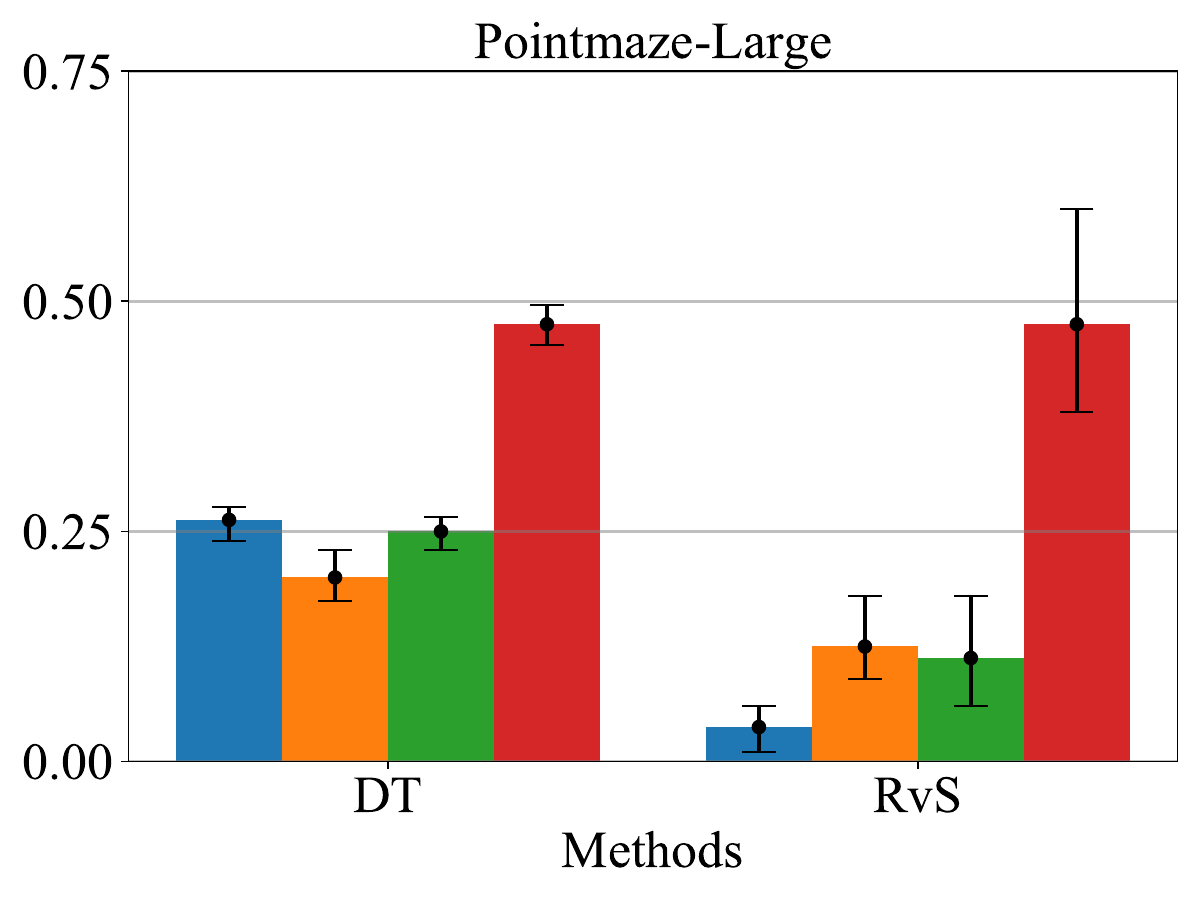}}
	\end{minipage}
    \vspace{-.25em}
    \caption{
     Performance of the original OCBC, as well as OCBC with corresponding goal data augmentation, compared to our SL method \textbf{GC\textit{Rein}SL} on the \texttt{Pointmaze} datasets from \citet{ghugare2024closing}.
     Error bars denote 95$\%$ bootstrap confidence intervals.
     \textbf{GC\textit{Rein}SL} not only improves the performance of DT and RvS in all tasks, but also outperforms exist goal data augmentation methods.
    }
    \vspace{-6pt}
    \label{fig:point goal results}
\end{figure*}
\begin{figure}[t]
    \vspace{-6pt}
    \centering
    \begin{minipage}{0.245\linewidth}
		\centerline{\includegraphics[width=0.7\textwidth]{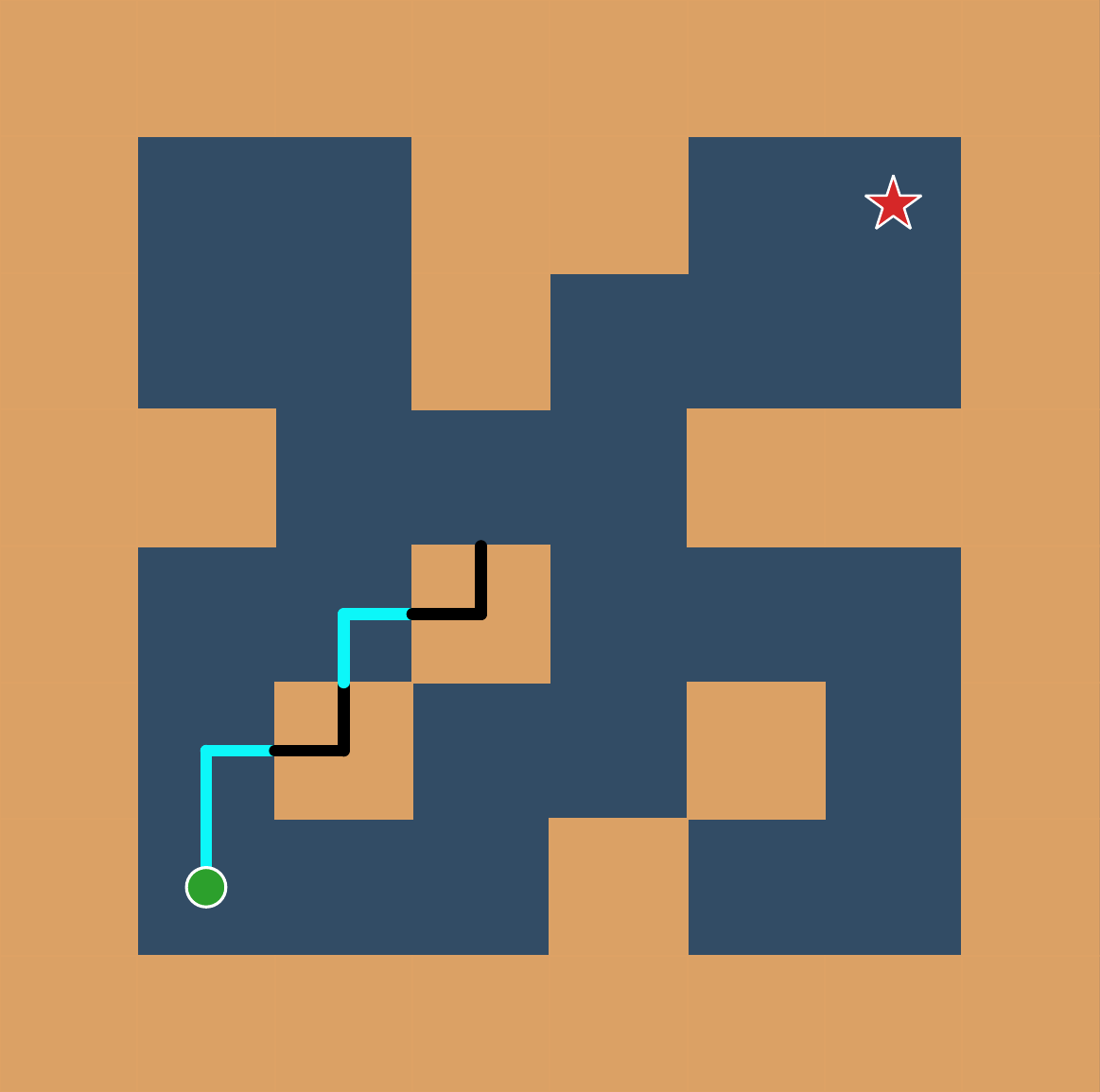}}
        \centerline{DT}
	\end{minipage}
    \begin{minipage}{0.245\linewidth}
		\centerline{\includegraphics[width=0.7\textwidth]{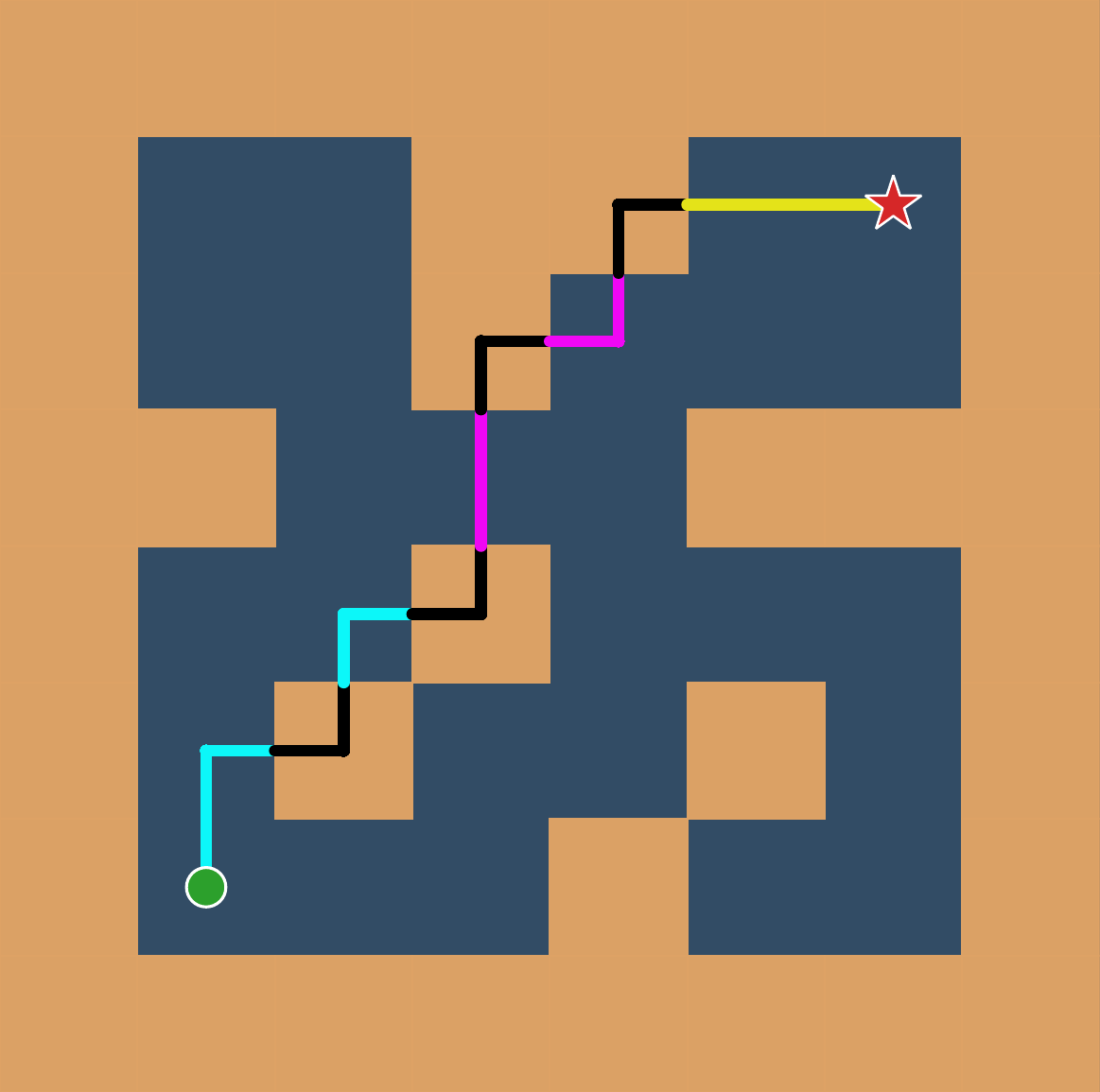}}
        \centerline{TGDA}
	\end{minipage}
	\begin{minipage}{0.245\linewidth}
		\centerline{\includegraphics[width=0.7\textwidth]{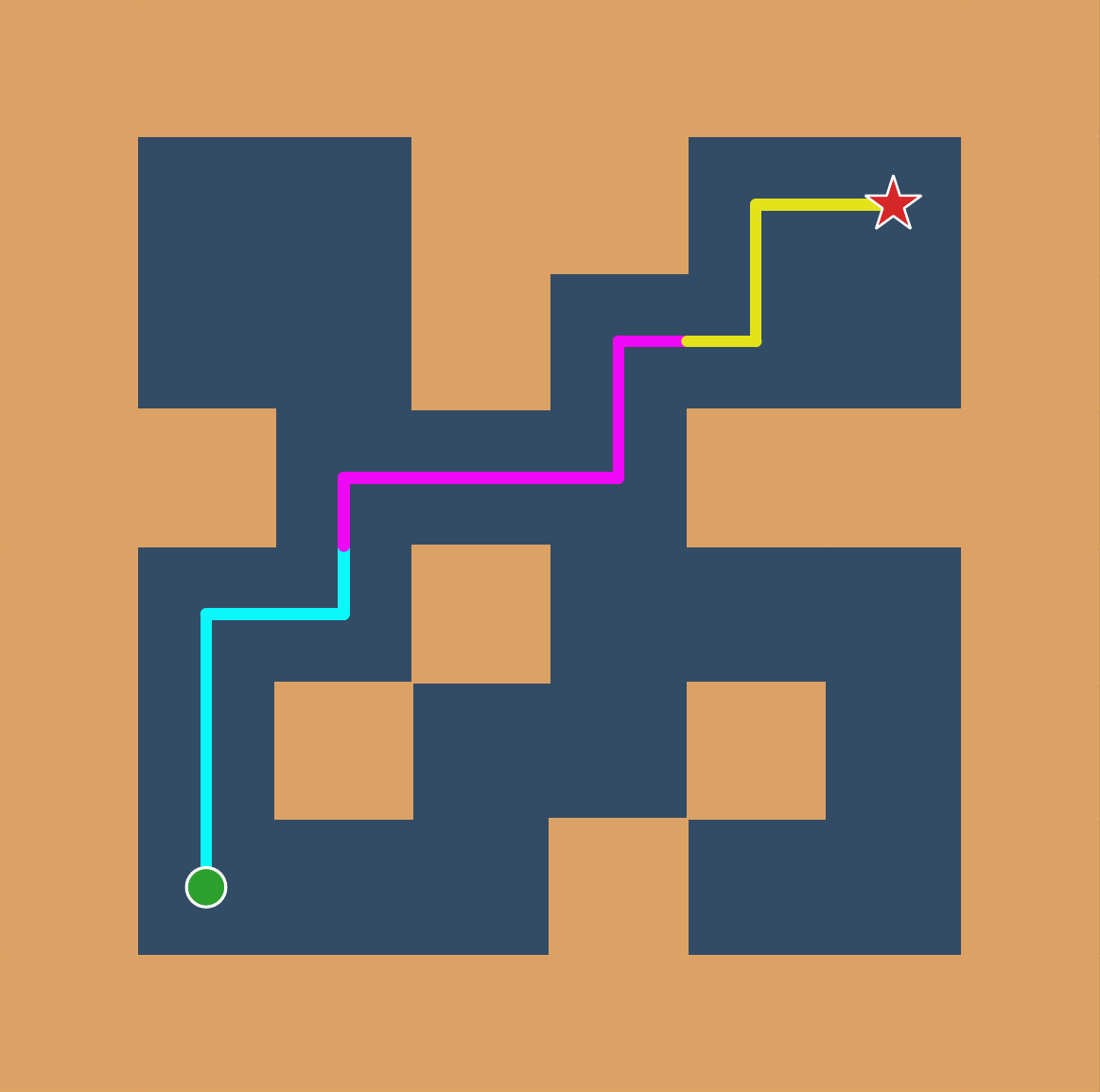}}
        \centerline{\textbf{GC\textit{Rein}SL}}
	\end{minipage}
    \begin{minipage}{0.245\linewidth}
		\centerline{\includegraphics[width=0.7\textwidth]{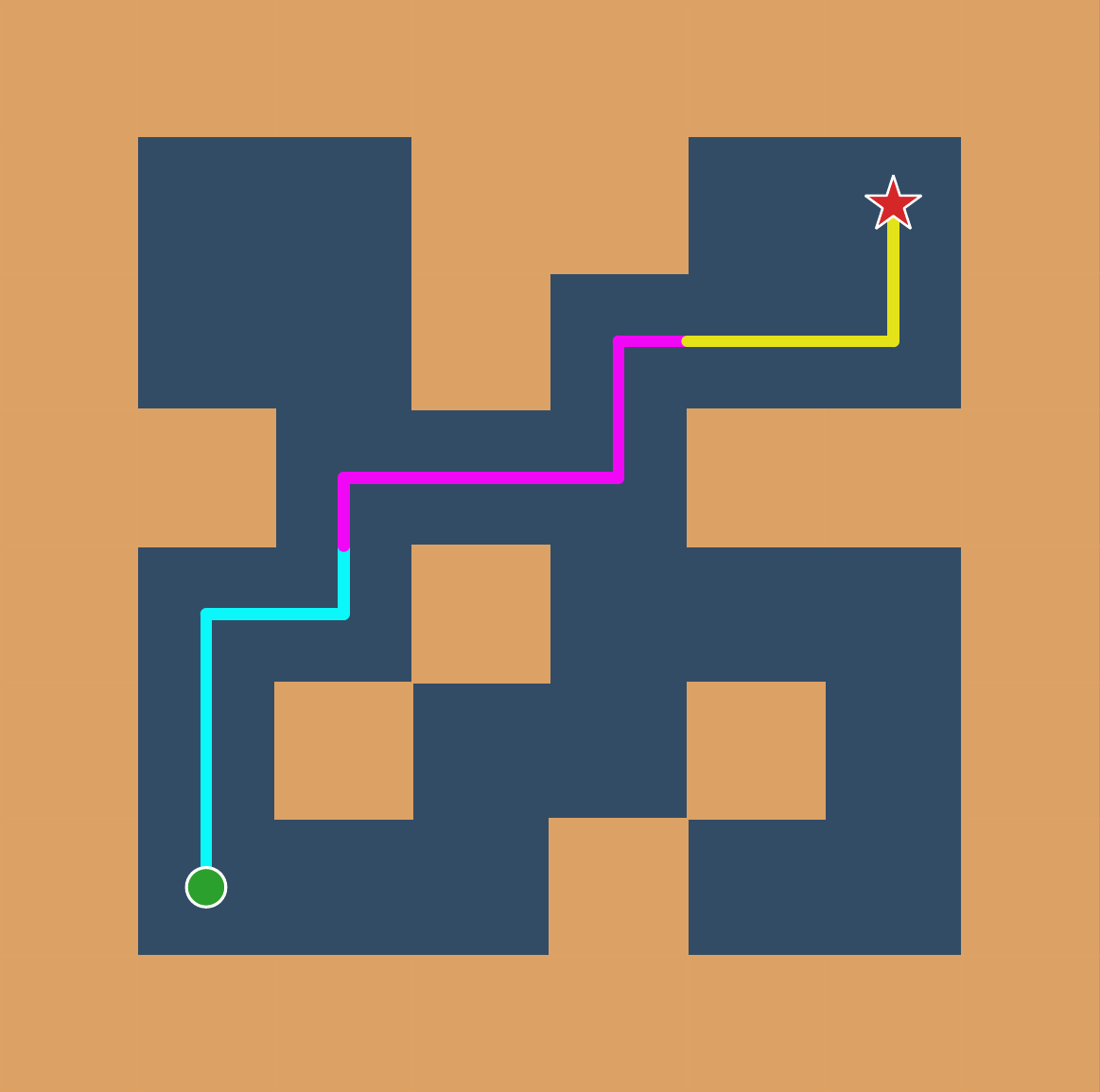}}
        \centerline{IQL}
	\end{minipage}
    \caption{
     Qualitative Comparison of DT, TGDA, \textbf{GC\textit{Rein}SL} for DT and IQL on \citet{ghugare2024closing} \texttt{Pointmaze-Medium} task. We observe that DT is unable to reach the specified goal (upper right) from start state (bottom left) and lacks stitching capability. Although TGDA can reach the specified goal, it frequently generates trajectories that cross walls, as it tends to prioritize OOD goals. In contrast, our GC\textit{Rein}SL address this issue, and the degree of stitching is comparable to that of IQL.
    }
    \vspace{-6pt}
    \label{fig:visual results}
\end{figure}
As shown in \cref{fig:point goal results}, it is evident that DT and RvS are struggle to possess stitching property, particularly in the \texttt{Pointmaze-Umaze} and \texttt{Pointmaze-Large} tasks, where their performance is notably poor. However, when $Q$-conditioned maximization is incorporated into the OCBC methods, performance improvements are observed across all tasks, albeit to varying degrees. This enhancement is attributed to the fact that \textbf{GC\textit{Rein}SL} allows for tackling unseen state-goal combination tasks during the inference phase, thereby improving the generalization and stitching capability of the models. Our \textbf{GC\textit{Rein}SL} consistently outperforms the other data augmentation approaches across all \texttt{Pointmaze} tasks, particularly in the more complex \texttt{Pointmaze-Medium} and \texttt{Pointmaze-Large} tasks. The qualitative comparison in 
\cref{fig:visual results} indicate that \textbf{GC\textit{Rein}SL}, while being SL-based, can effectively address long-horizon tasks that require trajectory stitching similar to the TD-based method IQL. 

\begin{figure*}[h]
    \vspace{-6pt}
    \centering
	\begin{minipage}{\linewidth}
		\centerline{\includegraphics[width=8.5cm]{results_pdf/results1.pdf}}
    \end{minipage}
    \begin{minipage}{0.32\linewidth}
		\centerline{\includegraphics[width=0.9\textwidth]{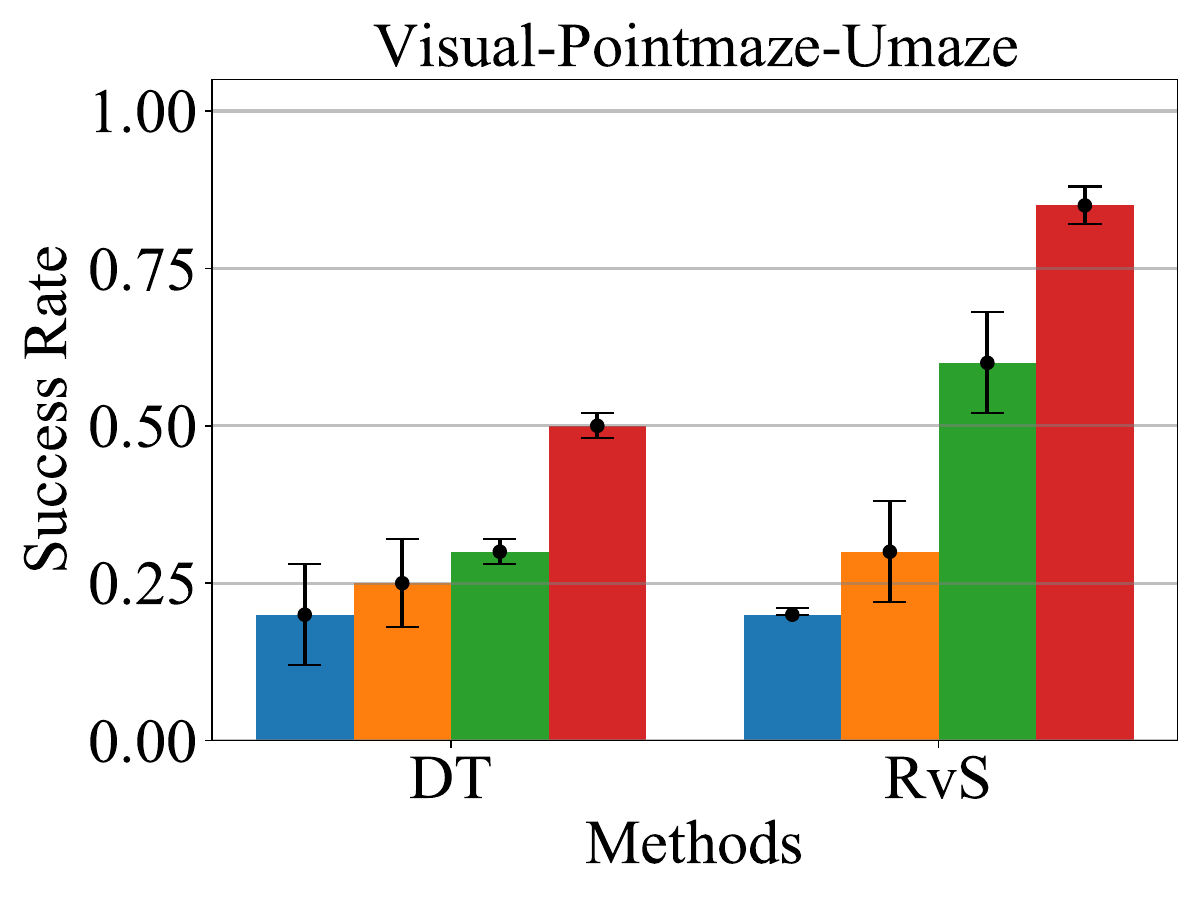}}
	\end{minipage}
    \begin{minipage}{0.32\linewidth}
		\centerline{\includegraphics[width=0.89\textwidth]{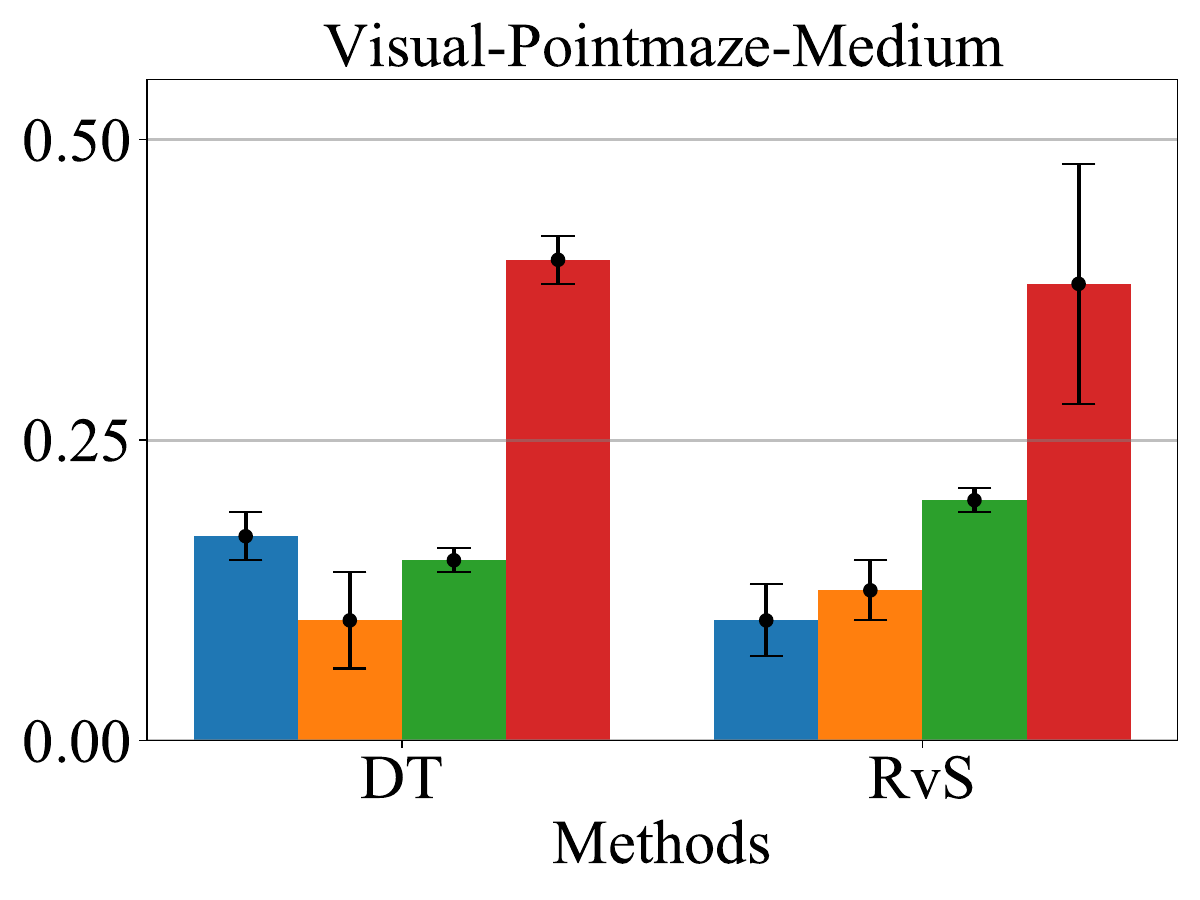}}
	\end{minipage}
	\begin{minipage}{0.32\linewidth}
		\centerline{\includegraphics[width=0.89\textwidth]{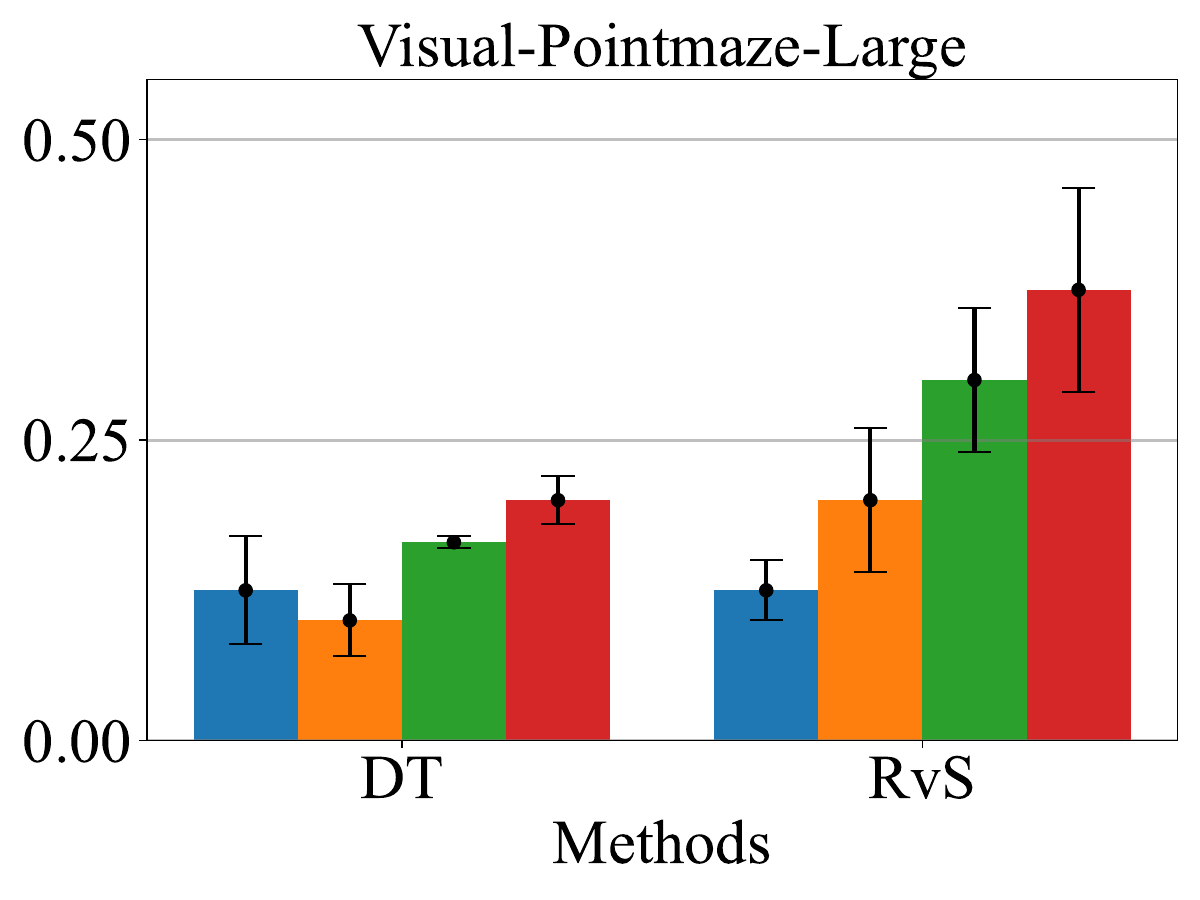}}
	\end{minipage}
    \caption{
     Performance of the original OCBC, as well as OCBC with corresponding goal data augmentation, compared to our SL method on the \texttt{Visual-Pointmaze} datasets from \citet{ghugare2024closing}.
     We use the final mean success rate as the report.  Error bars denote 95$\%$ bootstrap confidence intervals.
     \textbf{GC\textit{Rein}SL} not only improves the performance of DT and RvS in all tasks, but also outperforms existing goal data augmentation methods.
    }
    \vspace{-6pt}
    \label{fig:visual point goal results}
\end{figure*}
\subsection{Results in Higher-dimensional Visual Inputs} \label{sc:visual-input}
To evaluate the performance of our \textbf{GC\textit{Rein}SL} to tasks with higher-dimensional input observations, we implemented it on \texttt{Visual-Pointmaze} and \texttt{Antmaze} described in \citet{ghugare2024closing}.
As shown in Figure \ref{fig:visual point goal results}, the comparison between \textbf{GC\textit{Rein}SL} and OCBC, related state-of-the-art goal data augmentation methods in the \texttt{Visual-Pointmaze} dataset, demonstrates its scalability to visual observations. \textbf{GC\textit{Rein}SL} enhances the stitching performance of OCBC methods across all tasks, highlighting the
strength of SL methods in datasets with diverse state-goal distributions. It is noteworthy that SGDA exhibits the lowest robustness, performing even worse than the original DT on the \texttt{Visual-Pointmaze-Medium} and \texttt{Visual-Pointmaze-Large} dataset. This suggests that the random selection of goals may result in the inclusion of numerous low-quality goals, such as unreachable goals \citep{yang2023swapped}. In Appendix \ref{sc:visual-input}, we report the results on \texttt{Antmaze} datasets. We find that on all datasets, compared to other data augmentation methods, our \textbf{GC\textit{Rein}SL} (almost) always performs better than previous approaches, demonstrating that our method remains effective in the high-dimensional problem setting.

\begin{table*}[h]
    \vspace{-6pt}
    \centering
    \caption{The normalized best score on D4RL \citep{fu2020d4rl} \texttt{Antmaze-v2} datasets. The results come from its original Reinformer \citep{zhuang2024reinformer} paper except \textbf{GC\textit{Rein}SL}. The best result is \textbf{bold} and the \textcolor{blue}{blue} result means the best result among sequence modeling.}
    \resizebox{1.\textwidth}{!}{
    \begin{tabular}{c|cc|cccccc}
    \toprule
    \multicolumn{1}{c|}{\multirow{2}[4]{*}{\texttt{Antmaze-v2}}} & \multicolumn{2}{c|}{RL (Use TD)} & \multicolumn{6}{c}{Sequence Modeling (No TD)} \\
    \cmidrule{2-9}          & \multicolumn{1}{c}{CQL} & \multicolumn{1}{c|}{IQL} & \multicolumn{1}{c}{DT}  & \multicolumn{1}{c}{EDT} & \multicolumn{1}{c}{CGDT} & \multicolumn{1}{c}{Reinformer} & \multicolumn{1}{c}{QT (1-step)} & \multicolumn{1}{c}{\textbf{GC\textit{Rein}SL (ours)}} \\
    \midrule
    \multicolumn{1}{l|}{\texttt{umaze}} & \textbf{94.8 $\pm$ 0.8}   & 84.00 $\pm$ 4.1  & 64.5 $\pm$ 2.1   & 67.8$\pm$ 3.2   &71.0 & 84.4$\pm$2.7   &82.3$\pm$4.3 & \textcolor{blue}{85.1$\pm$5.3} \\
    \multicolumn{1}{l|}{\texttt{umaze-diverse}} & 53.8 $\pm$ 2.1   & \textbf{79.5 $\pm$ 3.4}  & 60.5 $\pm$ 2.3   & 58.3$\pm$ 1.9   &71.0 & 65.8$\pm$4.1   &80.8$\pm$5.1 & \textcolor{blue}{84.2$\pm$5.3} \\
    \multicolumn{1}{l|}{\texttt{medium-play}} & \textbf{80.5 $\pm$ 3.4}   & 78.5 $\pm$ 3.8  & 0.8 $\pm$ 0.4   & 0.0$\pm$ 0.0   &$\mathrm{/}$ & 13.2$\pm$6.1   &48.8$\pm$2.2  & \textcolor{blue}{49.0$\pm$3.5} \\
    \multicolumn{1}{l|}{\texttt{medium-diverse}} & 71.0 $\pm$ 4.5   & \textbf{83.5 $\pm$ 1.8}  & 0.5 $\pm$ 0.5   & 0.0$\pm$ 0.0   &$\mathrm{/}$ & 10.6$\pm$6.9   &49.2$\pm$2.4 & \textcolor{blue}{51.7$\pm$4.4} \\
    \multicolumn{1}{l|}{\texttt{large-play}} & 34.8 $\pm$ 5.9  & \textbf{53.5 $\pm$ 2.5}  & 0.0 $\pm$ 0.0  & 0.0$\pm$ 0.0  &$\mathrm{/}$ & 0.4 $\pm$0.5 &36.5$\pm$6.4 & \textcolor{blue}{38.2$\pm$1.8} \\
    \multicolumn{1}{l|}{\texttt{large-diverse}} & 36.3 $\pm$ 3.3  & \textbf{53.0 $\pm$ 3.00}  & 0.0 $\pm$ 0.0  & 0.0$\pm$ 0.0  &$\mathrm{/}$ & 0.4 $\pm$0.5 &30.2$\pm$6.2 & \textcolor{blue}{30.7$\pm$2.4} \\
    \cmidrule{1-9}    \textit{Total} & \textit{371.2} & \textit{432.0} & \textit{126.3}  &\textit{126.7} &$\mathrm{/}$ & \textit{174.8} &\textit{327.8} & \textcolor{blue}{\textit{338.9}} \\
    \bottomrule
    \end{tabular}}
    \vspace{-6pt}
    \label{tb:ant goal results} 
\end{table*}
\subsection{Return-conditioned RL Datasets Results}
We also extend our \textbf{GC\textit{Rein}SL} to return-conditioned RL (see \cref{appendix: return-conditioned rl} for detailed extensions) and compare it with advanced sequence modeling, as shown in \cref{tb:ant goal results}. From \cref{tb:ant goal results}, it is evident that in the majority of the \texttt{Antmaze-v2} datasets, particularly in the complex medium and large \texttt{Antmaze-v2} tasks, the \textbf{GC\textit{Rein}SL} approach demonstrates
superior performance, significantly closing the gap with TD-based methods such as CQL.
Compared to the two most closely related works EDT \citep{wu2023elastic} and Reinformer \citep{zhuang2024reinformer}, we utilize the estimated $Q$-value instead of their return-to-go \citep{chen2021decision}, which more accurately reflects the quality of actions during the stitching process \citep{wang2024critic,kim2024adaptive}.
\subsection{Ablation Study}\label{sc:ablation study}
\begin{wrapfigure}[15]{r}{0.7\linewidth}
    \vspace{-6pt}
    \centering
    \begin{minipage}{0.45\linewidth}
		\centerline{\includegraphics[width=\textwidth]{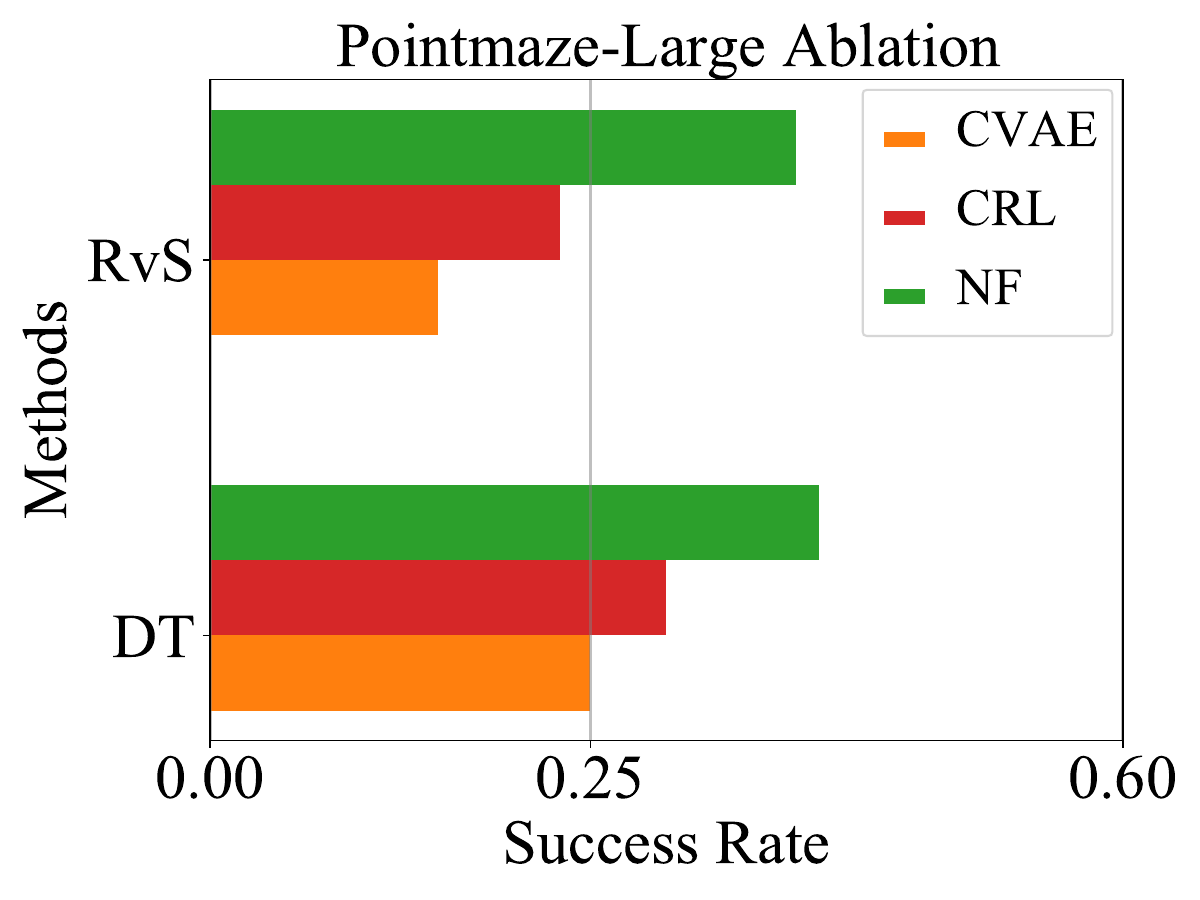}}
	\end{minipage}
    \begin{minipage}{0.45\linewidth}
		\centerline{\includegraphics[width=\textwidth]{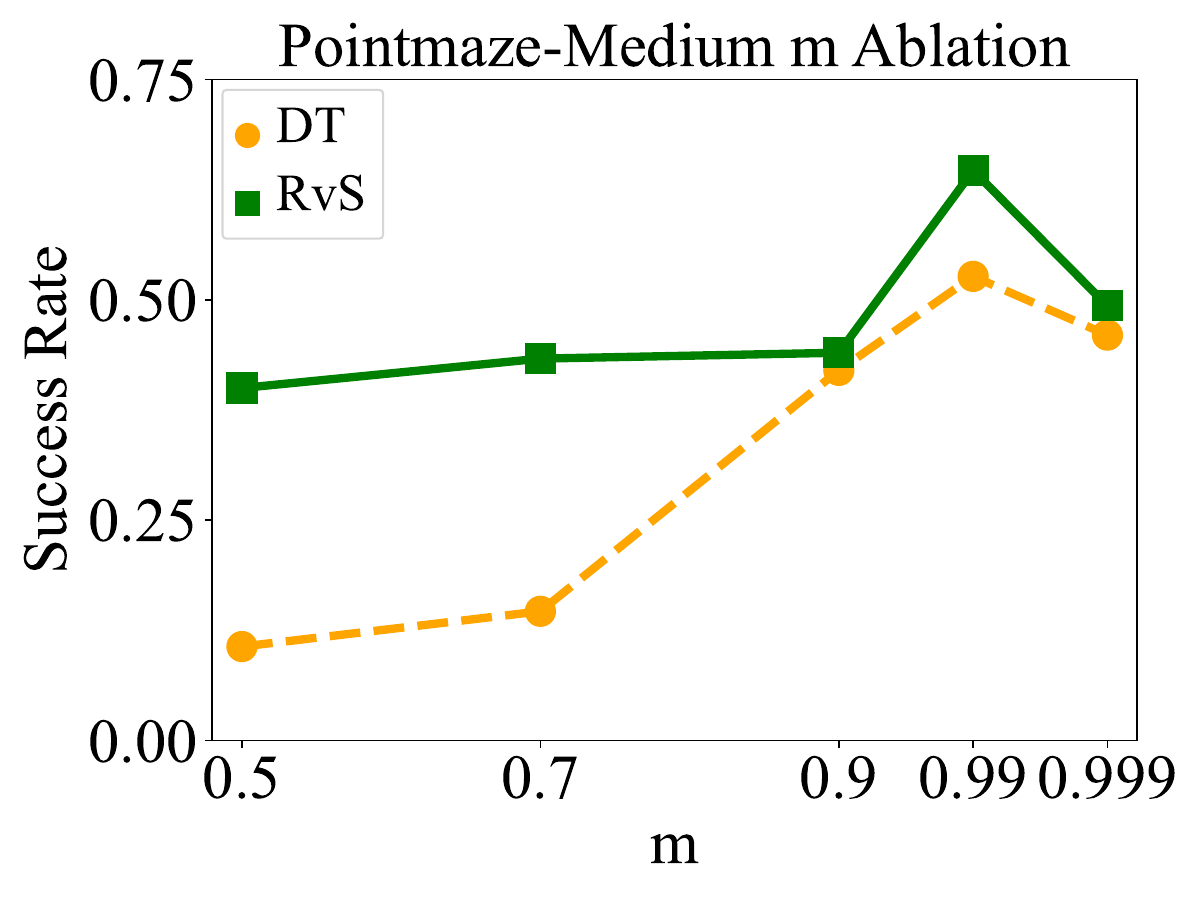}}
	\end{minipage}
    \caption{
    Ablation study of different probability estimators and $m$ in \citet{ghugare2024closing} datasets. \textbf{Left}: The performance on the \texttt{Pointmaze-Large} task. \textbf{Right}: The trend of last results as $m$ varies on \texttt{Pointmaze-Medium} task.
    }
    \vspace{-6pt}
    \label{fig:ablation results}
\end{wrapfigure}

In this section, we analyze the impact of different probability estimators and the value of $m$ in the $Q$-function loss (\cref{eq:total_loss}) on final performance. We select three distinct probabilistic model estimators—CVAE, CRL, and Normalizing Flows—for comparison. Previous work \citep{ghugare2025normalizing} in the literature has shown that Normalizing Flows provide more accurate estimates than the other two models. As demonstrated in the left panel of \cref{fig:ablation results}, \textbf{GC\textit{Rein}SL} is also influenced by the accuracy of the probability model estimator; the more accurate the estimate, the better the performance. These consistent findings across visual input tasks demonstrate that Normalizing Flows are not only highly effective for estimating multimodal goal distributions, but also represent the optimal approach for modeling goal-reaching probability.

As outlined in \cref{theorem:2}, as $m \rightarrow 1$, the learned $Q$-function asymptotically converges to the maximum $Q$-function within the offline distribution.
Given that a higher \textit{in-distribution} $Q$-function corresponds to improved action selection, we can infer that performance will improve as $m$ approaches 1. The experimental results presented in the right panel of \cref{fig:ablation results} are consistent with this theoretical prediction. However, larger values of $m$ do not consistently lead to more effective training or higher performance; in some cases, they may result in a performance decline. This could be attributed to overfitting to excessively large $Q$-values present in the offline dataset.
\section{Conclusion}
\label{sec:conclusion}
In this work, we introduce a $Q$-conditioned maximization supervised learning framework, embedding the maximized $Q$-value into SL-based methods (OCBC). To implement this framework, we propose the \textbf{GC\textit{Rein}SL} algorithm. Both theoretical analysis and experimental results demonstrate that \textbf{GC\textit{Rein}SL} significantly enhances the stitching capability of OCBC as well as sequence modeling methods while maintaining robustness. Future work could focus on developing more advanced OCBC architectures to further close the gap with TD learning.

\bibliographystyle{plainnat}
\bibliography{ref}

\setcounter{tocdepth}{-1}

\clearpage
\appendix
\setlength{\parindent}{0pt}
\renewcommand{\contentsname}{Contents of Appendix}
\renewcommand{\thesection}{\Alph{section}}

\addtocontents{toc}{\protect\setcounter{tocdepth}{3}} 

\begingroup
\hypersetup{linkcolor=black}
\begin{spacing}{1.5}
\tableofcontents 
\end{spacing}
\endgroup

\clearpage
\section{Proof of Theorem \ref{theorem:2}} \label{pf:2}
\begin{theorem}
We first define $\mathbf{SG} \dot= \left(s,g,a,Q^\beta\right)$.
For $m\in\left(0,1\right)$, if we denote $\mathbf{Q}^m\left(\mathbf{SG}\right) = \arg \min_{\hat{Q}} \mathcal{L}_{\hat{Q}}^m\left(\mathbf{SG}\right)$, then we have
\begin{align*}
    \lim_{m\rightarrow 1} \mathbf{Q}^m\left(\mathbf{SG}\right) = Q_{\text{max}}\,, \: \forall s, g\,,
\end{align*}
where $Q_{\text{max}} = \max_{\mathbf{a} \sim \mathcal{D}} Q^{\beta}\left(s,a,g\right)$ denotes the maximum $Q$-value with actions estimated from the offline dataset and $\mathcal{L}_{\hat{Q}}^m$ is define in \cref{eq:expectile regression}.
\end{theorem}
\paragraph{Proof} The proof primarily relies on the monotonicity property of $m$-expectile regression and employs a proof by contradiction.

Firstly, leveraging the monotonicity property of $m$-expectile regression \citep{newey1987asymmetric}, it follows that $\mathbf{Q}^{m_1} \leq \mathbf{Q}^{m_2}$ for $0<m_1<m_2<1$.

Secondly, for all $m \in (0, 1)$, it holds that $\mathbf{Q}^{m} \leq Q_{\text{max}}$. Assume there exists some $m_3$ such that $\mathbf{Q}^{m_3} > Q_{\text{max}}$. In this case, all $Q$-values from the offline dataset would satisfy $Q^\beta < \mathbf{Q}^{m_3}$. Consequently, the $Q$-function loss can be simplified given the same weight $1 - m_3$:
\begin{align*}
    \mathcal{L}^{m_3}_{\mathbf{Q}}&=\mathbb{E}\left[\left(1-m_3 \right)\left(Q^\beta - \mathbf{Q}^{m_3}\right)^2 \right] \\
    &> \mathbb{E}\left[\left(1-m_3 \right)\left(Q^\beta - Q_{\text{max}}\right)^2 \right].
\end{align*}
This inequality holds because $Q^\beta \leq Q_{\text{max}} < \mathbf{Q}^{m_3}$. However, this contradicts the fact that $\mathbf{Q}^{m_3}$ is derived by minimizing the $Q$-function loss. Therefore, the assumption is invalid, and we conclude that $\mathbf{Q}^{m} \leq Q_{\text{max}}$ is true. This proof step demonstrates that the predicted $Q$-function does not suffer from out-of-distribution (OOD) issues.

Finally, the convergence to this limit is a direct consequence of the properties of bounded and monotonically non-decreasing functions, thereby demonstrating the validity of the theorem.
\section{Extension in Return-conditioned RL} \label{appendix: return-conditioned rl}
\begin{wrapfigure}[18]{r}{0.4\linewidth}
    \centering
	\centerline{\includegraphics[width=0.4\textwidth]{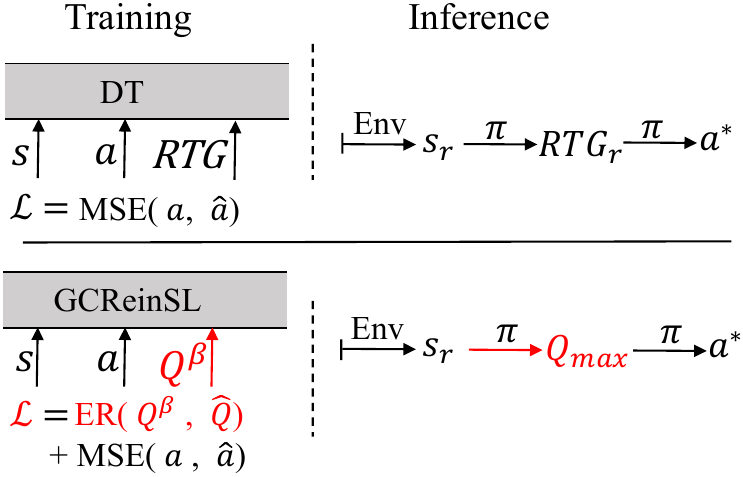}}
    \caption{
    \textbf{Left and Right at the Top:} DT. \textbf{Left and Right at the Bottom:} \textbf{GC\textit{Rein}SL}. $s$, $a$, RTG and $Q^\beta$ are come from offline data $\mathcal{D}$. 
    $s_r$ comes from environment. ER denotes Expectile Regression. The red section highlights the differences. 
    }
    \label{fig:dt comparasion results}
\end{wrapfigure}
To further clarify the differences between DT and our \textbf{GC\textit{Rein}SL} in return-conditioned RL, as well as the benefits of these changes, we first provide a comparison of the structure of DT and \textbf{GC\textit{Rein}SL} in \cref{fig:dt comparasion results}. We can observe that our \textbf{GC\textit{Rein}SL} replace the return-to-go (RTG) \citep{chen2021decision} conditioning with the estimated $Q^\beta$ during policy training, and employs expectile regression loss to obtain the maximized \textit{in-distribution} $Q$-value ${Q}_{max}$. The inference process determines the optimal action $a^{*}$ by considering both the given state and model predicted \textit{\textit{in-distribution}} maximum $Q$-value $Q$, rather than the \textit{arbitrarily selected} RTG $RTG_r$ in DT.

The primary benefit of the aforementioned changes stems from the learning of the $Q$-function, enabling the agent to obtain higher-quality actions more effectively during the stitching process \citep{kim2024adaptive}. Additionally, during training, our $Q^\beta$-conditioning effectively learns the mapping between the \textit{in-distribution} $Q$-value and the corresponding actions in the dataset.
In the inference phase, we condition our approach on the maximum $Q$-value supported by the dataset, thus eliminating the gap between training and inference while pursuing performance. Unlike DT, which learn the mapping between RTG  and action from the dataset during training but selects an \textit{arbitrary} RTG during the inference phase, whose appropriate can be suspicious. In \cref{sec:experiments} we experimentally compare our method and DT, and highlights the importance of an appropriate conditioning $Q$-value.

\section{\textbf{GC\textit{Rein}SL} Implementation Details} \label{app:alg}
\begin{figure}[h] 
\vspace{-6pt}
\centering
\centerline{\includegraphics[width=0.97\textwidth]{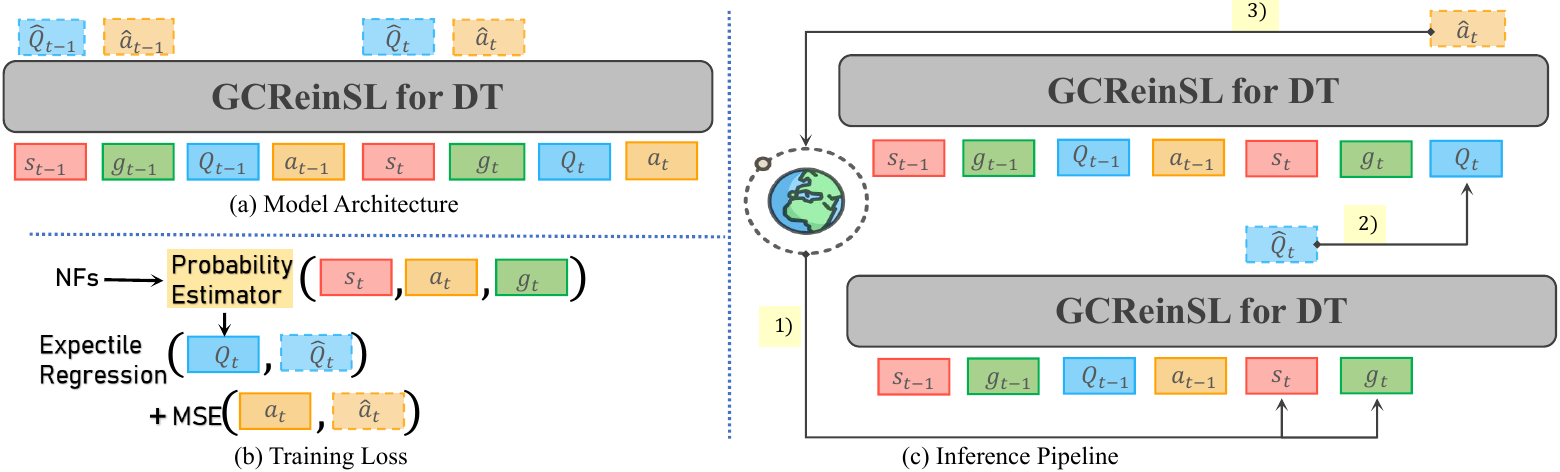}}
\caption{Overview of \textbf{GC\textit{Rein}SL} for DT:
(a) Model Architecture: The $Q$-function is the third inputs of \textbf{GC\textit{Rein}SL} for DT and the outputs contain $Q$-function and actions.
(b) Train Loss: As a $Q$-conditioned maximization sequence model, \textbf{GC\textit{Rein}SL} for DT not only maximizes the action likelihood but also maximizes $Q$-function by expectile regression. NFs denotes Normalizing Flows.
(c) Inference Pipeline: When inference, \textbf{GC\textit{Rein}SL} for DT first 1) gets state and goal from the environment to predict the \textit{in-distribution} maximum $Q$-function. Then 2) predicted \textit{in-distribution} max $Q$-function is concatenated with state and goal to predict the optimal action. Finally, 3) the environment executes the predicted action to $Q$-function the next state.} 
\vspace{-6pt}
\label{dt_overview}
\end{figure}

In this section we focus on the specific implementation of
\textbf{GC\textit{Rein}SL}, describing the architecture input and output, training, and inference procedures. Specifically, this section describes the training and inference pipeline using typical OCBC algorithm DT. Other supervised learning algorithms can be implemented in a similar manner. The overall structure of \textbf{GC\textit{Rein}SL} for DT is depicted in \cref{dt_overview}, with RvS being similar, differing only in terms of its architecture.
\subsection{Implementation of \textbf{GC\textit{Rein}SL} for DT}
\paragraph{Model Architecture} 
To accommodate the $Q$-conditioned maximization for DT \citep{chen2021decision}, which predicts the maximum $Q$-value and utilizes it as a condition to guide the generation of optimal actions, we position $Q$-value between state and goal. 
In detail, the input token sequence of \textbf{GC\textit{Rein}SL} for DT and corresponding output tokens are summarized as follows:
\begin{align*}
    \textbf{Input: \ }&\left<\ \cdots,s^{\left(n\right)}_t, g^{\left(n\right)}_t, Q^{\left(n\right)}_t, a^{\left(n\right)}_t\right> \\
    \textbf{Output: \ }&\quad \quad \left<\ \hat{Q}^{\left(n\right)}_t, \hat{a}^{\left(n\right)}_t,\Box\ \right>
\end{align*}
$s^{\left(n\right)}_t$, $g^{\left(n\right)}_t$, $Q^{\left(n\right)}_t$ and $a^{\left(n\right)}_t$ represent individual tokens within the DT.
When predicting the $\hat{Q}_t^{\left(n\right)}$, the model takes the current state $s_t^{\left(n\right)}$ and previous $K$ timesteps tokens $\left<s,g,Q,a\right>_{t-K}^{\left(n\right)}=\big(s^{\left(n\right)}_{t-K+1}, g^{\left(n\right)}_{t-K+1},Q^{\left(n\right)}_{t-K+1}, a^{\left(n\right)}_{t-K+1},\cdots,s^{\left(n\right)}_{t-1},g^{\left(n\right)}_{t-1},$ $Q^{\left(n\right)}_{t-1}, a^{\left(n\right)}_{t-1}\big) $ into consideration.
For the sake of simplicity, $\mathbf{SG}^{\left(n\right)}_{t-K}$ denotes the input $\left[\left<s,g,Q,a\right>^{\left(n\right)}_{t-K};s^{\left(n\right)}_t,g^{\left(n\right)}_t\right]$.
While the action prediction $\hat{a}_t$ is based on $\left(\mathbf{SG}^{\left(n\right)}_{t-K}, \mathbf{Q}^{\left(n\right)}_{t-K}\right) = \left[\left<s,g,Q,a\right>^{\left(n\right)}_{t-K};s^{\left(n\right)}_t,g^{\left(n\right)}_t,Q^{\left(n\right)}_t\right]$.
The $\Box$ means that this predicted token neither participates in training nor inference.
At timestep $t$, different type of tokens are embedded by different linear layers and fed into the transformers \citep{vaswani2017attention} together.
The output $Q$-function $\hat{Q}^{\left(n\right)}_t$ is processed by a linear layer.
\paragraph{Training Loss} Since the model predicts both $\hat{Q}_t$ and $\hat{a}_t$, its training loss consists of a $Q$-function loss and an action loss. For the action loss, we adopt the MSE loss function of DT and simultaneously adjust the order of tokens:
\begin{equation}\label{action_loss}
\mathcal{L}_{\text{a}} = \mathbb{E}_{t,n}\bigg[a_t^{\left(n\right)} -  \pi_{\theta}\left(\mathbf{SG}^{\left(n\right)}_{t-K}, \mathbf{Q}^{\left(n\right)}_{t-K}\right)\bigg]^2.
\end{equation}
The $Q$-function loss is the expectile regression with the parameter $m$:
\begin{align}\label{Q-function_loss}
    \mathcal{L}^{m}_{\text{Q}}=\mathbb{E}_{t, n}&\left[\left|m-\mathbbm{1} \left( \Delta Q < 0\right) \right|\Delta Q^2\right], 
    \text{with \ } \Delta Q = Q_t^{(n)} - \pi_{\theta}\left(\mathbf{SG}^{\left(n\right)}_{t-K}\right).
\end{align}
We use the same weight for these two loss functions and therefore the total loss is $\mathcal{L}_{\text{a}} + \mathcal{L}^{m}_{Q}$.

\paragraph{Inference Pipeline}
For each timestep $t$, the action is the last token, which means the predicted action is affected by state from the environment and the $Q$-function.
The $Q$-function of the trajectories output by the sequence model exhibits a positive correlation with the initial conditioned $Q$-function \citep{chen2021decision,zheng2022online}. 
That is, within a certain range, higher initial $Q$-function typically lead to better actions.
In classical $Q$-learning \citep{mnih2015human}, the optimal value function $Q^*$ can derive the optimal action $a^*$ given the current state. 
In the context of sequence modeling, we also assume that the maximum $Q$-value is required to output the optimal actions. 
The inference pipeline of the \textbf{GC\textit{Rein}SL} is summarized as follows:
\begin{align}
    \overset{\text{\textcolor{blue}{Env}}}{\longmapsto} \left(s_0,g_0\right) \xrightarrow{\pi_{\theta}} \hat{Q}_0 \xrightarrow{\pi_{\theta}} a_0 \xrightarrow{\text{\textcolor{blue}{Env}}} \left(s_1,g_1\right) \xrightarrow{\pi_{\theta}} \hat{Q}_1 \xrightarrow{\pi_{\theta}} a_1 \rightarrow \cdots
\end{align}
Specially, the environment initializes the state-goal pair $\left(s_0,g_0\right)$ and then the sequence model $\pi_{\theta}$ predicts the maximum $Q$-value $\hat{Q}_0$ given current state-goal pair $\left(s_0,g_0\right)$.
Concatenating $\hat{Q}_0$ with $\left(s_0,g_0\right)$, $\pi_{\theta}$ guarantees the output of the optimal action $a_0$.
It is important to note that  $\left(s_0,g_0\right)$ may be derived from a cross-trajectory. In this case, our $\pi_{\theta}$ can still output the optimal action.
Then the environment transitions to the next state $s_1$ and receive the new goal $g_1$.
Repeat the above steps until the trajectory comes to an end.
\subsection{\textbf{GC\textit{Rein}SL} Algorithm for DT}\label{ap:2.1}
\begin{algorithm}[h]
    \caption{\textbf{GC\textit{Rein}SL} for DT}
    \begin{algorithmic}[1]\label{training_inference}
    \STATE {\bfseries Input:} offline dataset $\mathcal{D}$, sequence modeling $\pi_{\theta}$
    \STATE  Initialize Normalizing Flows with parameters $\psi$
    \STATE \textbf{Function} Normalizing Flows Training
    \STATE \quad Sample minibatch of transitions from offline dataset $\mathcal{D}$: $\left(s, a, g\right) \sim \mathcal{D}$
    \STATE \quad Update $\psi$ maximizing \cref{eqn:nfs}
    \STATE \textcolor{blue}{//Training Procedure}
    \FOR{ sample $ \left<\ \cdots,s_{t}, g_{t}, a_{t}\ \right>$ from $\mathcal{D}$}
        \STATE Get $Q_t$ with probability estimator with \cref{eqn:nfq}
        \STATE Get $\hat{Q}_t, \hat{a}_t$ with sequence modeling $\pi_{\theta}$: $\hat{Q}_t, \hat{a}_t=\pi_{\theta}\left(\cdots,s_{t}, g_{t}, a_{t}, Q_t\right)$
        \STATE Calculate total loss $\mathcal{L}_{\text{a}} + \mathcal{L}_{\text{Q}}^m$ by \cref{action_loss} and \cref{Q-function_loss}, and take a gradient descent step on $\nabla_{\theta}\left(\mathcal{L}_{\text{a}} + \mathcal{L}_{Q}^m\right)$
    \ENDFOR
    \STATE \textcolor{blue}{//Inference Pipeline}
    \STATE {\bfseries Input:} sequence modeling $\pi_{\theta}$, environment $\text{Env}$
    \STATE $s_0 = \text{Env}.reset(\ )$ and $t=0$
    \REPEAT
    \STATE Predict maximum $Q$-function $\textcolor{blue}{\hat{Q}_t}=\pi_{\theta}\left(\cdots,s_{t}, g_{t}, \Box, \Box\ \right)$
    \STATE Predict optimal action $\hat{a}_t=\pi_{\theta}\left(\cdots,s_{t}, g_{t}, \textcolor{blue}{\hat{Q}_{t}}, \Box \right)$
    \STATE $s_{t+1},r_t= \text{Env}.step(\hat{a}_t)$ and $t=t+1$
    \UNTIL{done}
\end{algorithmic}
\end{algorithm}
\subsection{Implementation of \textbf{GC\textit{Rein}SL} for RvS} \label{ap:rvs imp details}
\paragraph{Architecture} 
To accommodate the $Q$-conditioned maximization for RvS \citep{emmons2021rvs}, which also predicts the maximum $Q$-function and utilizes it as a condition to guide the generation of optimal actions. Unlike \textbf{GC\textit{Rein}SL} for DT, we construct a actor model for predicting actions and a value model $v_{\phi}$ for predicting $V$-function \footnote{In this paper, we do not make a strict distinction between the $V$-function and the $Q$-function, treating their meanings as equivalent.}. 
In detail, the input of \textbf{GC\textit{Rein}SL} for RvS and corresponding output are summarized as follows:
\begin{align*}
    \textbf{Input: \ }&s_t, g_t, Q_t(s_t, a_t, g_t) \\
    \textbf{Value Model Output: \ }&\hat{V}_t(s_t, g_t) \\
    \textbf{Actor Model Output: \ }&\hat{a}_t\left(s_t, g_t, \hat{V}_t(s_t, g_t)\right)
\end{align*}
When predicting the $\hat{V}_t$, the value model takes the current state $s_t$ and desired goal $g_t$.
For action $\hat{a}_t$, we adopt a actor model that incorporates $V$-values for inference.
\paragraph{Training Procedure and Inference Pipeline} Like \textbf{GC\textit{Rein}SL} for DT, the total loss function is also composed of $Q$ ($V$)-function loss and action loss, and the form is the same.
At each step of the inference pipeline, the value model outputs the maximum $V$-value for the input state-goal pair, and then the actor model outputs the corresponding action.
Note that in this state-goal pair, the state and the goal are treated as distinct elements.
In the context of RvS, we also assume that the maximum $V$-value are required to output the optimal actions. The training procedure is similar to that of \textbf{GC\textit{Rein}SL} for DT, with the key distinction that the prediction of $V$-value is generated by a value model.
The inference pipeline of the \textbf{GC\textit{Rein}SL} is summarized as follows:
\begin{align}
    \overset{\text{\textcolor{blue}{Env}}}{\longmapsto} \left(s_0,g_0\right) \xrightarrow{v_{\phi}} \hat{V}_0 \xrightarrow{\pi_{\theta}} a_0 \xrightarrow{\text{\textcolor{blue}{Env}}} \left(s_1,g_1\right) \xrightarrow{v_{\phi}} \hat{V}_1 \xrightarrow{\pi_{\theta}} a_1 \rightarrow \cdots
\end{align}
Specially, the environment initializes the state-goal pair $\left(s_0,g_0\right)$, and then the value model $v_{\phi}$ predicts the maximum $V$-value $\hat{V}_0$ given current state-goal pair.
Concatenating $\hat{V}_0$ with $\left(s_0,g_0\right)$, $\pi_{\theta}$ can output the optimal action $a_0$.
Then the environment transitions to the next state $s_1$ and the desired goal $g_1$.

\subsection{\textbf{GC\textit{Rein}SL} Algorithm for RvS}\label{ap:2.2}
\begin{algorithm}[H]
\caption{\textbf{GC\textit{Rein}SL} for RvS}
\begin{algorithmic}[1]\label{inference_2}
    \STATE {\bfseries Input:} offline dataset $\mathcal{D}$, actor model $\pi_{\theta}$, value model $v_{\phi}$
    \STATE Normalizing Flows training is similar to \textbf{GC\textit{Rein}SL} for DT.
    \STATE \textcolor{blue}{//Training Procedure}
    \FOR{ sample $ \left<\ \cdots,s_{t}, g_{t}, a_{t}\ \right>$ from $\mathcal{D}$}
        \STATE Get $Q_t$ with probability estimator with \cref{eqn:nfq}
        \STATE Predict maximum $V$-value $\textcolor{blue}{\hat{V}_t}=v_{\phi}\left(s_{t},g_{t}\right)$
        \STATE Predict optimal action $\hat{a}_t=\pi_{\theta}\left(s_{t},g_{t},\textcolor{blue}{\hat{V}_{t}}\right)$
        \STATE The calculation of the total loss is also the same as in \textbf{GC\textit{Rein}SL} for DT.
    \ENDFOR
    \STATE \textcolor{blue}{//Inference Pipeline}
    \STATE {\bfseries Input:} value model $v_{\phi}$, actor model $\pi_{\theta}$, environment $\text{Env}$
    \STATE $s_0 = \text{Env}.reset(\ )$ and $t=0$
    \REPEAT
    \STATE Predict maximum $V$-function $\textcolor{blue}{\hat{V}_t}=v_{\phi}\left(s_{t},g_{t}\right)$
    \STATE Predict optimal action $\hat{a}_t=\pi_{\theta}\left(s_{t},g_{t},\textcolor{blue}{\hat{V}_{t}}\right)$
    \STATE $s_{t+1},r_t= \text{Env}.step(\hat{a}_t)$ and $t=t+1$
    \UNTIL{done}
\end{algorithmic}
\end{algorithm}
\section{Baseline Details}\label{sc:baseline details}
We compare our approach with a wide variety of baselines, including goal data augmentation based stitching methods, sequence modeling and TD-based RL methods.

Particularly, we include the following methods:
\begin{itemize}

\item For goal data augmentation methods, we include SGDA
\citep{yang2023swapped} and TGDA \citep{ghugare2024closing}.
SGDA proposes a method that randomly choose augmented goals from different trajectories.
TGDA employs $k$-means \citep{lloyd1982least} to cluster the goal and certain states into a group, and samples goals from later stages of these state trajectories as augmented goals. We employ these two goal data augmentation methods in conjunction with DT and RvS as baseline comparisons;

\item For sequence modeling methods, we include DT \citep{chen2021decision}, EDT \citep{wu2023elastic}, CGDT \citep{wang2024critic}, Reinformer \citep{zhuang2024reinformer} and QT (1-step) \citep{hu2024q}. 
DT is a classic sequence modeling method that utilizes a Transformer architecture to model and reproduce sequences from demonstrations,
integrating a goal-conditioned policy to convert Offline RL into a supervised learning task. Despite
its competitive performance in Offline RL tasks, the DT falls short in achieving trajectory stitching \citep{brandfonbrener2022does}.
EDT is a variant of DT that lies in its ability to determine the optimal history length to promote trajectory stitching. But it does not incorporates the RL objective that
maximizes returns to enhance the model \citep{zhuang2024reinformer} and its stitching capabilities are limited \citep{kim2024adaptive}.
Reinformer is similar to our work; however, it exhibits limited stitching capabilities due to the absence of $Q$-value, resulting in a significant performance gap compared to TD-based RL methods.
QT introduces Q-value regularization to optimize action selection on top of DT and excels in handling long time horizons and sparse reward tasks. We selected the 1-step variant of QT, which is most closely aligned with our approach, for comparison and denote it as QT (1-step).

\item For TD-based RL methods, we include CQL~\citep{kumar2020conservative} and IQL~\citep{kostrikov2021offline}.
CQL and IQL are classical offline RL methods that utilize dynamic programming. This trick endows them with stitching properties \citep{cheikhi2023statistical,ghugare2024closing}.

\end{itemize}
\section{Experiment Details} \label{ap:exp details}
In this section we provide offline datasets details as well as implementation details used for all the
algorithms in our experiments – DT, RvS, Normalizing Flows, and \textbf{GC\textit{Rein}SL}.
\subsection{Offline Datasets} \label{sc:app_dataset}
\begin{figure}[h]
\vspace{-6pt}
\begin{center}
    \includegraphics[width=0.28\textwidth,height=3.3cm]{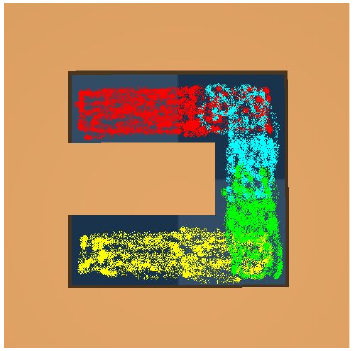}\label{fig:sub1}\hspace{0.3cm}
    \includegraphics[width=0.28\textwidth,height=3.3cm]{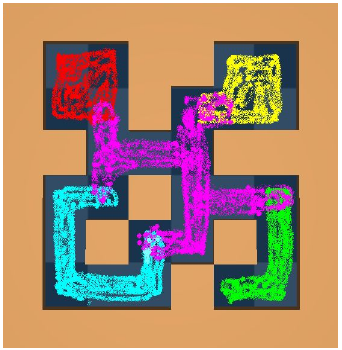}\label{fig:sub2}\hspace{0.3cm} 
    \includegraphics[width=0.28\textwidth,height=3.3cm]{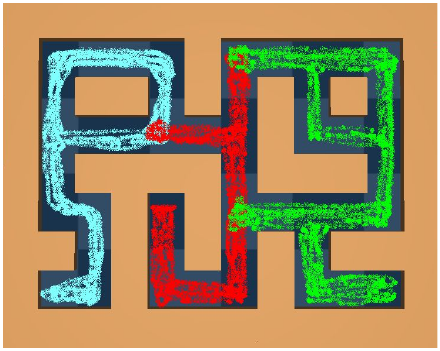}\label{fig:sub3}\\
\end{center}
    \text{~~~~~~~~~~~~~~~~~~~~~~~~~Umaze~~~~~~~~~~~~~~~~~~~~~~~~~~~~~~~~~~Medium~~~~~~~~~~~~~~~~~~~~~~~~~~~~~~~~~~~~~~Large}
  \caption{Goal-conditioned RL datasets from \citet{ghugare2024closing}: Different colors represent the navigation regions of various data collection policies. During data collection, these policies navigate between randomly selected state-goal pairs within their respective navigation regions. These visualizations pertain to the \texttt{Pointmaze}, with similar patterns observed in the \texttt{Antmaze} datasets.}
  \vspace{-6pt}
  \label{fig:point-weak-dataset}
\end{figure}

\paragraph{Goal-conditioned RL} 
We utilize the \texttt{Pointmaze} , \texttt{Visual-Pointmaze} and \texttt{Antmaze} datasets in \citet{ghugare2024closing}. As described in \cref{sec:experiments}, both offline datasets contain $10^6$ transitions and are specifically constructed to evaluate trajectory stitching in a combinatorial setting (see \cref{fig:point-weak-dataset}). In the \texttt{Pointmaze} dataset, the task involves controlling a ball with two degrees of freedom by applying forces along the Cartesian x and y axes. By contrast, the \texttt{Antmaze} dataset features a 3D ant agent, provided by the Farama Foundation~\citep{towers2023gymnasium}. The \texttt{Pointmaze} and \texttt{Visual-Pointmaze} were collected using a PID controller, while the \texttt{Antmaze} datasets were generated using a pre-trained policy from D4RL \citep{fu2020d4rl}. Visual representations of the various \texttt{Pointmaze} configurations can be found in \cref{fig:point-weak-dataset}.

\begin{figure}[h]
\vspace{-6pt}
\begin{center}
    \includegraphics[width=0.28\textwidth,height=3.3cm]{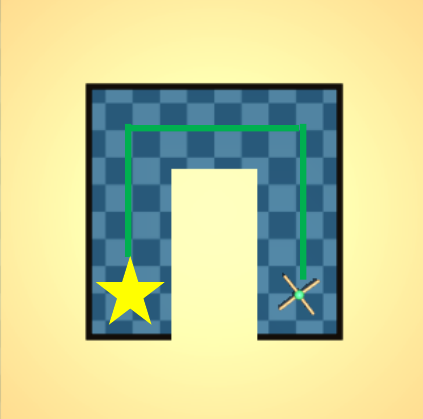}\label{fig:sub4}\hspace{0.3cm}
    \includegraphics[width=0.28\textwidth,height=3.3cm]{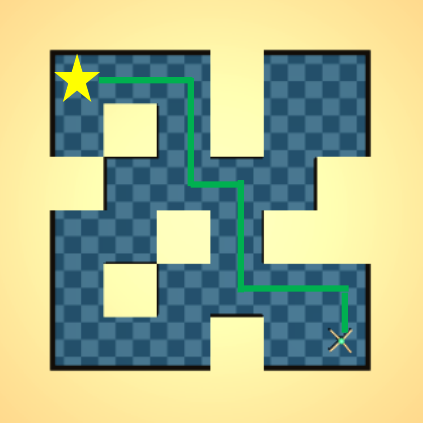}\label{fig:sub5}\hspace{0.3cm} 
    \includegraphics[width=0.28\textwidth,height=3.3cm]{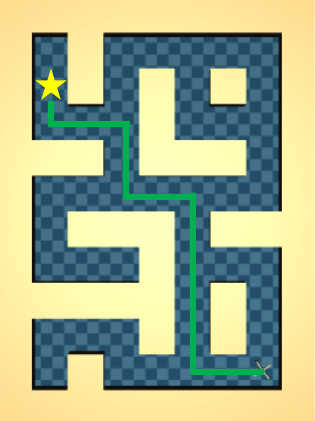}\label{fig:sub6}\\
\end{center}
    \text{~~~~~~~~~~~~~~~~~~~~~~~~~Umaze~~~~~~~~~~~~~~~~~~~~~~~~~~~~~~~~~~Medium~~~~~~~~~~~~~~~~~~~~~~~~~~~~~~~~~~~~~~Large}
  \caption{Return-conditioned RL Datasets from \citet{fu2020d4rl}: The \texttt{AntMaze-v2} datasets involve controlling an 8-DoF quadruped to navigate towards a specified goal state. This benchmark requires value propagation to effectively stitch together sub-optimal trajectories from the collected data.}\label{fig:Antmaze}
  \vspace{-6pt}
  \label{gym}
\end{figure}
\paragraph{Return-conditioned RL}
In the experiments comparing with related sequence modeling approaches, we follow the methodology outlined in \citet{zhuang2024reinformer} to construct the \texttt{AntMaze-v2} datasets using D4RL \citep{fu2020d4rl}, which also contain $10^6$ transitions (see \cref{fig:Antmaze}).
These \texttt{AntMaze-v2} datasets are characterized by sparse rewards, where $r = 1$ is awarded upon reaching the goal. The umaze, medium, and large datasets all lack complete trajectories from the starting point to the desired goal, necessitating that the algorithm reconstructs the desired trajectory by stitching together incomplete or failed segments.

\subsection{Implementation Details} \label{ap:imp details}
We ran all our experiments on NVIDIA RTX 8000 GPUs with 48GB of memory within an
internal cluster. In goal-conditioned RL, we use the default configurations of DT and RvS as described in \citet{ghugare2024closing}, with some values modified. In goal-conditioned RL, we use the default configurations of DT in \citet{zhuang2024reinformer}. The architecture and training process of the Normalizing Flows are identical to those described in
\citet{ghugare2025normalizing}. 

Our \textbf{GC\textit{Rein}SL} for DT implementation draws inspiration from and references the following three repositories:
\begin{itemize}
    \item TGDA: \url{https://github.com/RajGhugare19/stitching-is-}\\ \url{combinatorial-generalisation};
    \item Normalizing Flows: \url{https://github.com/Princeton-RL/}\\ \url{normalising-flows-4-reinforcement-learning};
    \item Reinformer: \url{https://github.com/Dragon-Zhuang/Reinformer}.
\end{itemize}
The state tokens, goal tokens, $Q$-function tokens and action tokens are first processed by different linear layers. 
Then these tokens are fed into the decoder layer to obtain the embedding.
Here the decoder layer is a lightweight implementation from Reinformer \citep{zhuang2024reinformer}.
The context length for the decoder layer is denoted as $K$. 
Our \textbf{GC\textit{Rein}SL} for RvS implementation is similar to the idea of \textbf{GC\textit{Rein}SL} for DT,
but it is divided into value networks and policy networks.
The value network outputs the expected $V$-function from state $s$ to goal $g$. This expected V-function, along with the state 
$s$ and goal $g$, is then used as input to the policy network.
We employed both the AdamW \citep{loshchilov2017decoupled} and Adam \citep{2014Adam} optimizers to optimize the total loss for DT and RvS, respectively, in alignment with the methods outlined in their original papers. The hyperparameter of $Q$-function loss is denoted as $m$.
\section{Hyperparameters} \label{ap:hy}
In this section, we will provide a detailed description of parameter settings in our experiments. The hyperparameters of SGDA \citep{yang2023swapped} and TGDA \citep{ghugare2024closing} remain consistent with their original settings. For fair comparison, our method still sets the same \textbf{data augmentation probability} of 0.5 as theirs. The default number of training steps is 50000, with a learning rate of 0.001.
With these default settings, if the training score continues to rise, we would consider increasing the number of training steps or doubling the learning rate.
For some datasets, 50000 steps may cause overfitting and less training steps are better. The hyperparameters of \textbf{GC\textit{Rein}SL} for DT in various datasets are presented in the tables below. 
In all tables, the arrows indicate the directional change in the corresponding values for RvS.
\subsection{Hyperparameter $m$}
The hyperparameter $m$ is crucially related to the $Q$-function loss and is one of our primary focuses for tuning. 
We explore values within the range of $m = [0.7, 0.9, 0.99, 0.999]$. 
When $m=0.5$, the expectile loss function will degenerate into MSE loss, which means the model is unable to output a maximized $Q$-function.
So we do not take $m=0.5$ into consideration.
We observe that performance is generally lower at $m=0.9$ compared to others except \texttt{Pointmaze-Umaze}.
Only \texttt{Pointmaze-Large} adopt the parameter $m=0.999$ while $m=0.99$ are generally better than $m=0.999$ on other datasets. The detailed hyperparameter selection of $m$ is summarized in the following \cref{tb:hyperparameter m}:

\begin{table*}[h]
    \centering
    \caption{Hyperparameters $m$ of $Q$-function loss on different datasets. }
    \vspace{-6pt}
    \resizebox{\linewidth}{!}{
    \begin{tabular}{l|r||l|r}
    \toprule
    \textbf{Dataset}                   & \textbf{$m$} & \texttt{Antmaze-umaze-v2}           & 0.9 \\ \cline{1-2}
    \texttt{(Visual) Pointmaze-Umaze}        & (0.9) $0.9 \rightarrow 0.99$          & \texttt{Antmaze-umaze-diverse-v2}          & 0.99 \\
    \texttt{(Visual) Pointmaze-Medium}        & (0.99) 0.99        & \texttt{Antmaze-medium-play-v2}          & 0.99 \\
    \texttt{(Visual) Pointmaze-Large} & ($0.9 \rightarrow 0.99$) $0.99 \rightarrow 0.999$          & \texttt{Antmaze-medium-diverse-v2}           & 0.99 \\ 
    \texttt{Antmaze-Umaze} & 0.99          & \texttt{Antmaze-large-play-v2}       & 0.99 \\
    \texttt{Antmaze-Medium/Large}             & 0.99        & \texttt{Antmaze-large-diverse-v2}        & 0.99  \\
    \bottomrule
    \end{tabular}}
    \label{tb:hyperparameter m}
\end{table*}
\subsection{Context Length $K$}
The context length $K$ is another key hyperparameter in \textbf{GC\textit{Rein}SL} for DT, and we conduct a parameter search across the values $K=[2, 5, 10, 20]$. 
The maximum value is $20$ because the default context length for DT \citep{chen2021decision} is $20$. 
The minimum is $2$, which corresponds to the shortest sequence length (setting $K=1$ would no longer constitute sequence learning). 
Overall, we found that $K=10$ and $K=20$ lead to more stable learning and better performance on \citet{ghugare2024closing} \texttt{Pointmaze} and \texttt{Antmaze} datasets. 
Conversely, a smaller context length is preferable on D4RL \texttt{Antmaze-v2} dataset.
The parameter $K$ has been summarized as follow \cref{tb:hyperparameter K}:

\begin{table*}[h]
    \centering    
    \caption{Context length $K$ on different datasets. }
    \vspace{-6pt}
    \begin{tabular}{l|r||l|r}
    \toprule
    \textbf{Dataset}                   & \textbf{$K$} & \texttt{Antmaze-umaze-v2}           & 2 \\ \cline{1-2}
    \texttt{(Visual) Pointmaze-Umaze}        & (10) 10          & \texttt{Antmaze-umaze-diverse-v2}          & 2 \\ 
    \texttt{(Visual) Pointmaze-Medium}        & (20) 10          & \texttt{Antmaze-medium-play-v2}          & 3 \\ 
    \texttt{(Visual) Pointmaze-Large} & (10) 5          & \texttt{Antmaze-medium-diverse-v2}           & 2 \\ 
    \texttt{Antmaze-Umaze} & 20          & \texttt{Antmaze-large-play-v2}       & 3 \\
    \texttt{Antmaze-Medium/Large}             & 20        & \texttt{Antmaze-large-diverse-v2}        & 2  \\
    \bottomrule
    \end{tabular}
    \label{tb:hyperparameter K}
\end{table*}
\section{Additional Results}
\label{appendix:additional-results}
This section evaluates the resilience of \textbf{GC\textit{Rein}SL} across several factors, including the average probability of improvement, visual-inputs results, the capability of Normalizing Flows to accurately estimate goal probabilities, the qualitative comparison, and training curves on goal-conditioned datasets from \citet{ghugare2024closing}. Due to space constraints, not all of these variations are discussed in the main body of this study. The details are provided below.

\begin{figure*}[t]
    \centering
    
    \begin{subfigure}{0.45\linewidth}
        \centering
        \includegraphics[width=\textwidth]{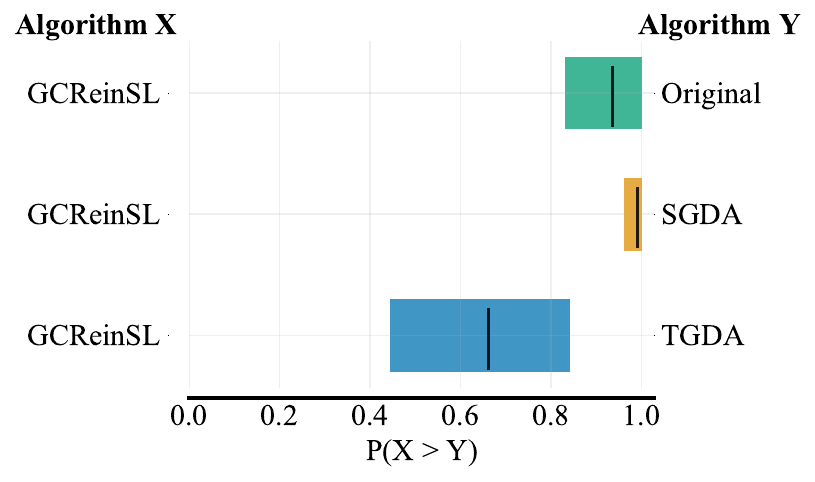}
        \caption{\textbf{GC\textit{Rein}SL} for RvS in \texttt{Pointmaze}}
    \end{subfigure}
    \hfill
    \begin{subfigure}{0.45\linewidth}
        \centering
        \includegraphics[width=\textwidth]{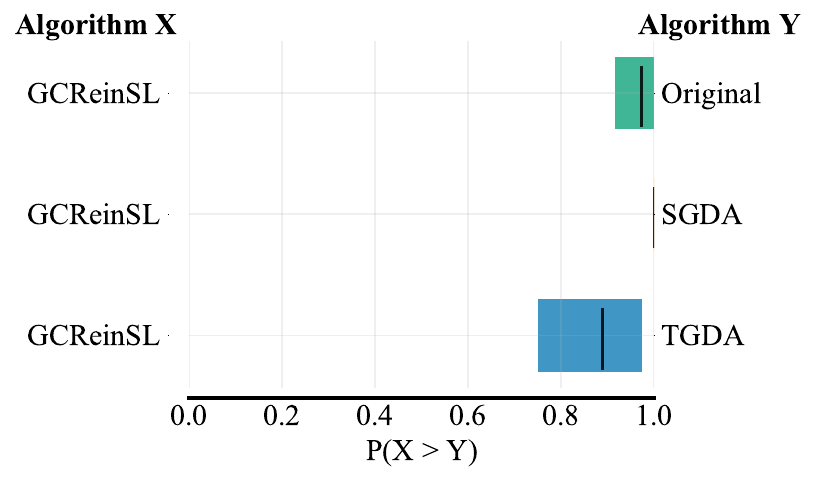}
        \caption{\textbf{GC\textit{Rein}SL} for DT in \texttt{Pointmaze}}
    \end{subfigure}
    
    \vspace{1em}
    
    \begin{subfigure}{0.45\linewidth}
        \centering
        \includegraphics[width=\textwidth]{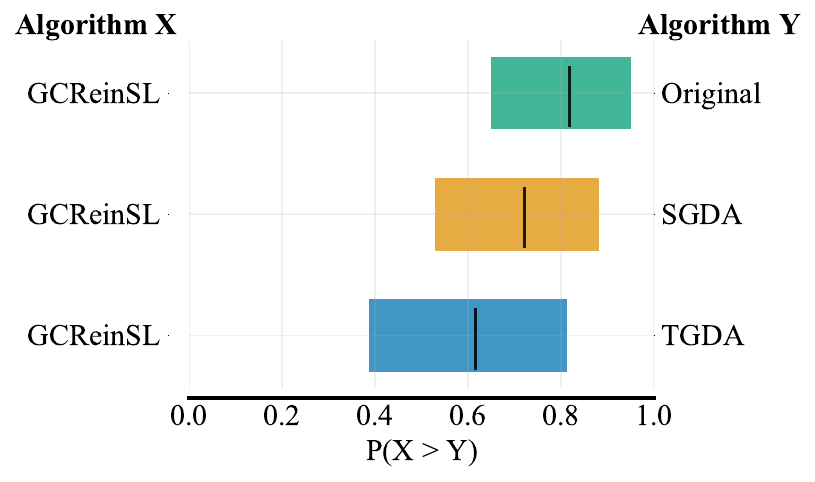}
        \caption{\textbf{GC\textit{Rein}SL} for RvS in \texttt{Antmaze}}
    \end{subfigure}
    \hfill
    \begin{subfigure}{0.45\linewidth}
        \centering
        \includegraphics[width=\textwidth]{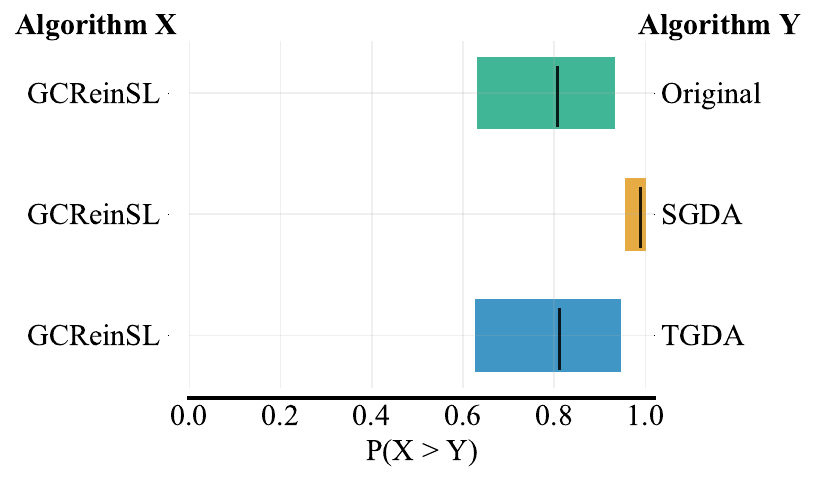}
        \caption{\textbf{GC\textit{Rein}SL} for DT in \texttt{Antmaze}}
    \end{subfigure}
    
    \caption{
        Average probability of improvement on offline (a) (b) \texttt{Pointmaze} and (c) (d) \texttt{Antmaze} datasets. Each figure shows the probability of improvement of \textbf{GC\textit{Rein}SL} compared to original or other data augmentation methods. The interval estimates are based on stratified bootstrap with independent sampling with 2000 bootstrap re-samples.
    }
    \label{fig:avg_prob_improvement}
\end{figure*}
\subsection{Average Probability of Improvement}
In this subsection, we adopt the average probability of improvement \citep{agarwal2021deep}, a robust metric to measure how likely it is for one algorithm to outperform another on a randomly selected task. The results are reported in \cref{fig:avg_prob_improvement}. As shown in the results, \textbf{GC\textit{Rein}SL} robustly outperforms other data augmentation baselines on the \texttt{Pointmaze} datasets. For instance, \textbf{GC\textit{Rein}SL} for DT is $98\%$ better than original DT method and $100\%$ better than SGDA. On the complex \texttt{Antmaze} datasets, the probability trend of outperforming the baselines is consistent, whether for \textbf{GC\textit{Rein}SL} for DT or \textbf{GC\textit{Rein}SL} for RvS. Note that the most two effective and robust algorithms on both \texttt{Pointmaze} and \texttt{Antmaze} datasets are \textbf{GC\textit{Rein}SL} and TGDA, which are specifically designed for trajectory stitching. Comparing the two algorithms, \textbf{GC\textit{Rein}SL} outperforms TGDA with a average probability of $77.5\%$ on the \texttt{Pointmaze} datasets and $62\%$ on the \texttt{Antmaze} datasets.

\begin{figure*}[t]
    
    \centering
	\begin{minipage}{\linewidth}
		\centerline{\includegraphics[width=8.5cm]{results_pdf/results1.pdf}}
    \end{minipage}
    \begin{minipage}{0.32\linewidth}
		\centerline{\includegraphics[width=\textwidth]{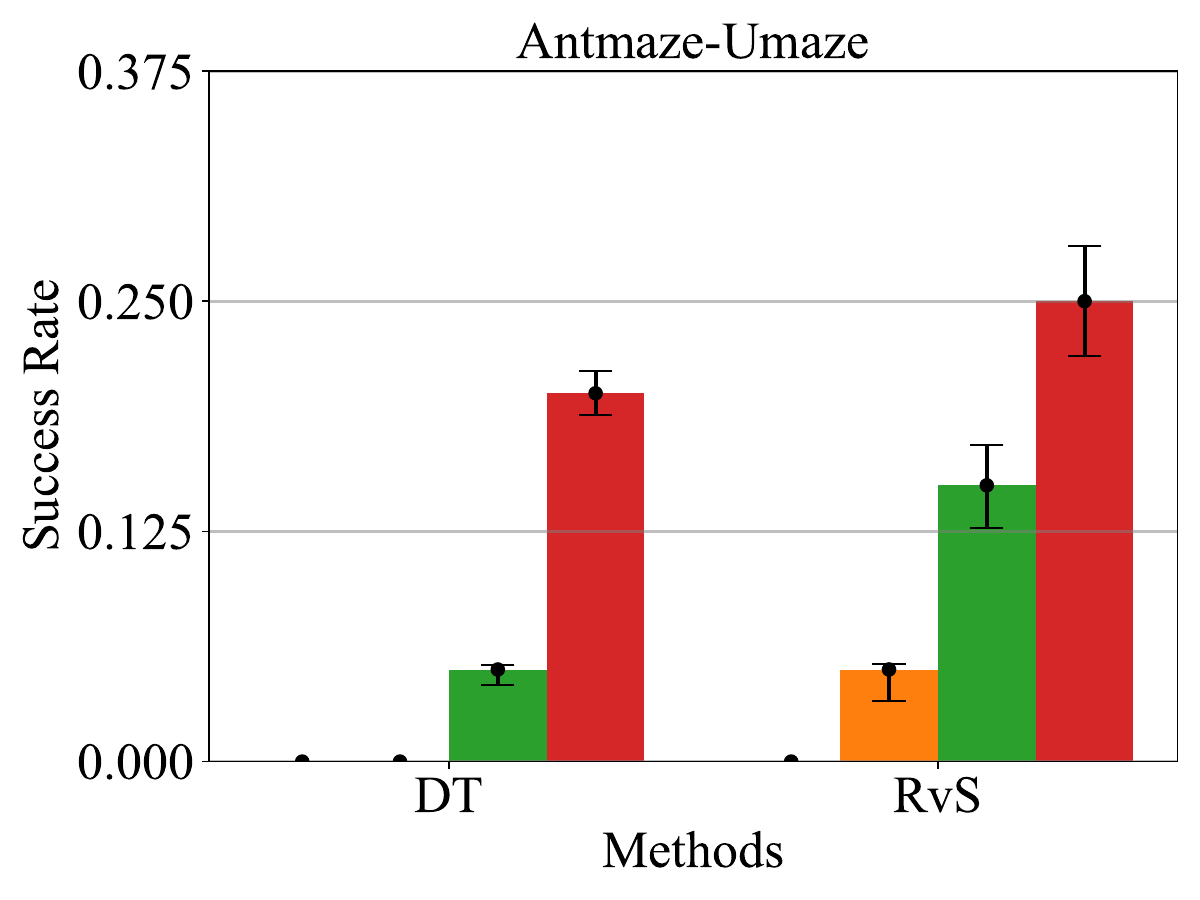}}
	\end{minipage}
    \begin{minipage}{0.32\linewidth}
		\centerline{\includegraphics[width=0.99\textwidth]{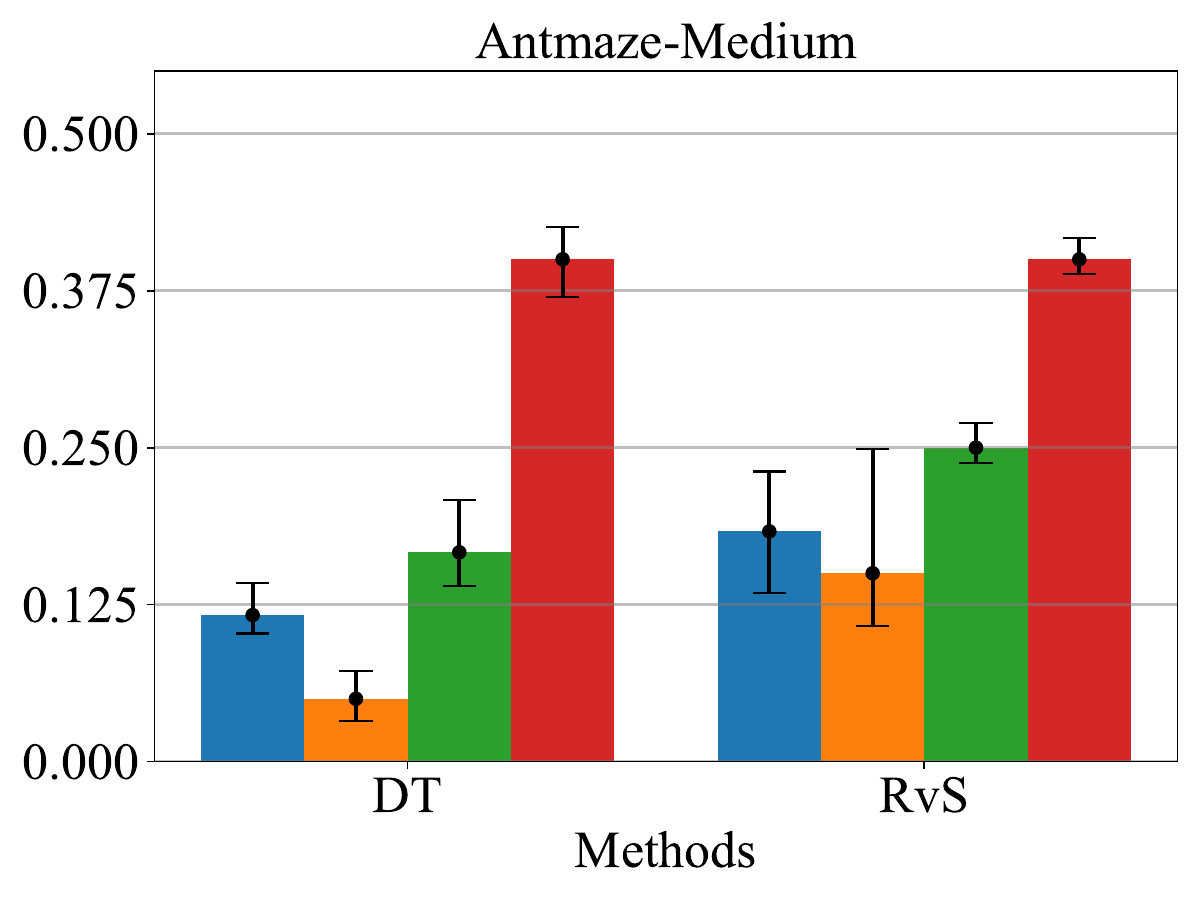}}
	\end{minipage}
	\begin{minipage}{0.32\linewidth}
		\centerline{\includegraphics[width=0.99\textwidth]{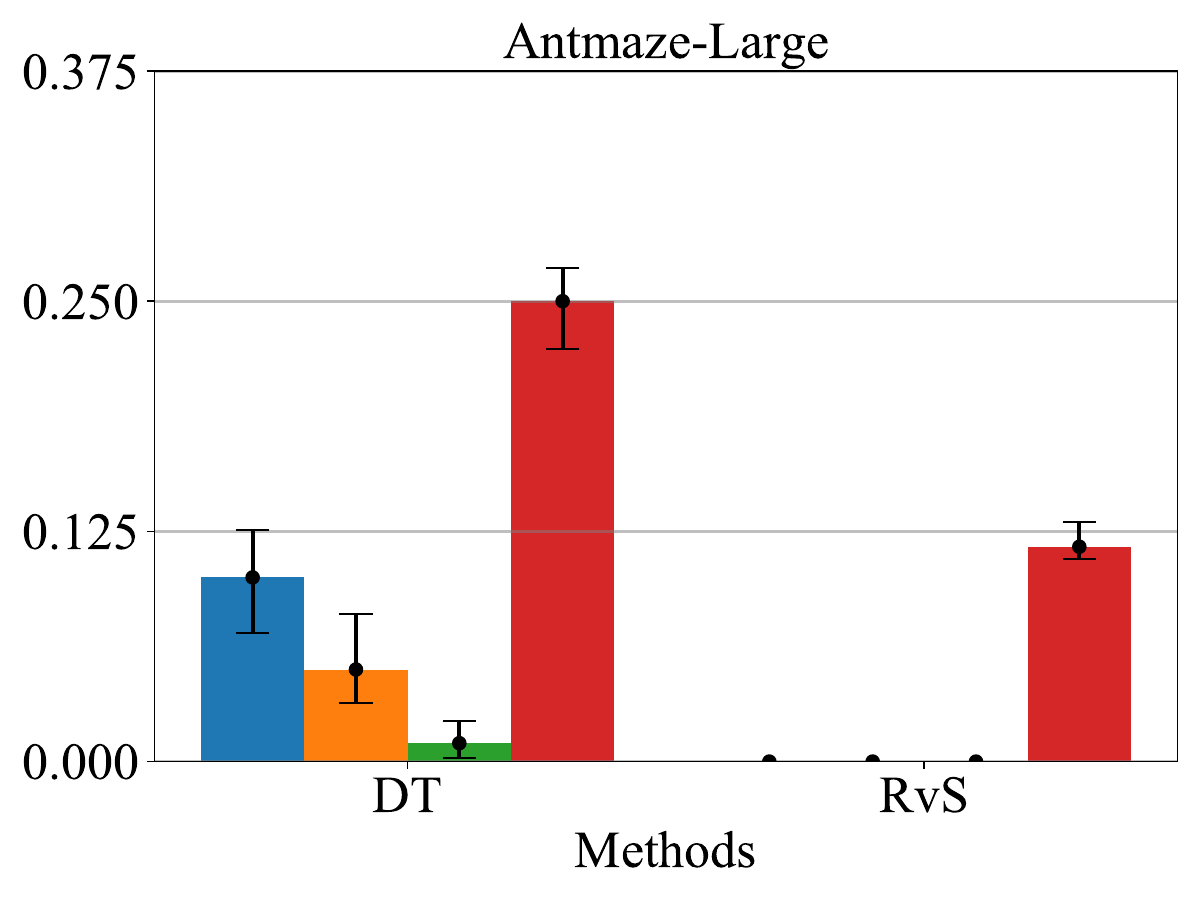}}
	\end{minipage}
    \caption{Performance on high-dimensional \citet{ghugare2024closing} \texttt{Antmaze} datasets.
    \textbf{GC\textit{Rein}SL} can consistently improve the performance of OCBC and surpass goal data augmentation methods on all high-dimensional \texttt{Antmaze} datasets. 
    Error bars denote 95$\%$ bootstrap confidence intervals.
    We demonstrate that through the learning and utilization of maximum in-distribution $Q$-value, \textbf{GC\textit{Rein}SL} enhances the stitching capability of OCBC.
    }
    
    \label{ap:ant goal results}
\end{figure*}
\subsection{Results in \textit{Antmaze} Datasets} \label{sc:visual-input}
In Figure \ref{ap:ant goal results}, we observe that \textbf{GC\textit{Rein}SL} improves the performance of DT and RvS across all \texttt{Antmaze} datasets, with particularly notable improvements on the medium and large datasets.
\subsection{Evaluating the Capability of Normalizing Flows to Accurately Estimate Goal-reaching Probability} \label{ap:evaluate nf}
In this section, we validate the accuracy of the Normalizing Flows's estimation of the discounted future state distribution by implementing the computation method outlined in \citet{eysenbach2020c} within a tabular setting. 
It is important to note that here we are solely validating the accuracy of the Normalizing Flows in estimating the discounted future state distribution, which is unrelated to the actual implementation of the Normalizing Flows in our \textbf{GC\textit{Rein}SL} framework.

Specifically, we compute the true discounted future state distribution in a modified GridWorld environment example and evaluate the estimation error by comparing it against the true distribution. We also compare the predictions of CVAE\citep{sohn2015learning}, C-learning \citep{eysenbach2020c} and CRL\citep{eysenbach2022contrastive} with the true future state density.
First, we introduce the modified GridWorld environment used in this experiment. This environment is characterized by 
stochastic dynamics and a continuous state space, such that the true $Q$-function for the indicator reward is zero.
Specifically, the environment has a size of $5 \times 5$, where the agent observes a noisy version of its current state. More precisely, when the agent is located at position $(i, j)$, it observes the state $(i + \epsilon_i, j + \epsilon_j)$, where $\epsilon_i, \epsilon_j \sim \text{Unif}[-0.5, 0.5]$.
Note that the observation uniquely identifies the agent's position, so there is no partial observability. Similar to \citet{eysenbach2020c}, we analytically compute the exact future state density function by first determining the future state density of the underlying GridWorld, noting that the density is uniform within each cell. We generated a tabular policy by sampling from a Dirichlet (1) distribution, and sampled 100 trajectories of length 100 from this policy for Normalizing Flows training.

\begin{figure*}[h]
  \centering
  \begin{minipage}{0.32\linewidth}
		\centerline{\includegraphics[width=\textwidth]{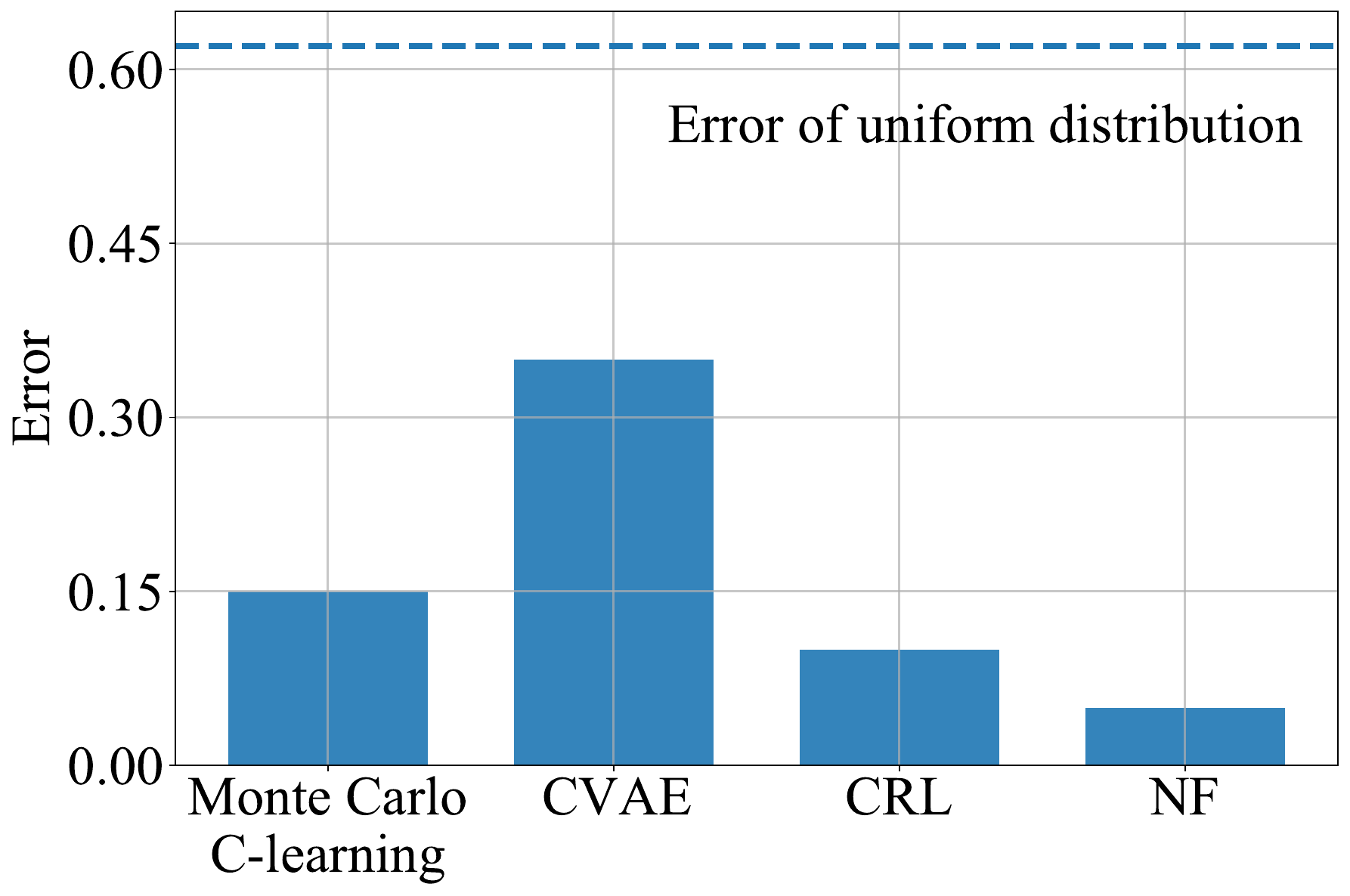}}
  \end{minipage}
  \begin{minipage}{0.32\linewidth}
		\centerline{\includegraphics[width=0.94\textwidth]{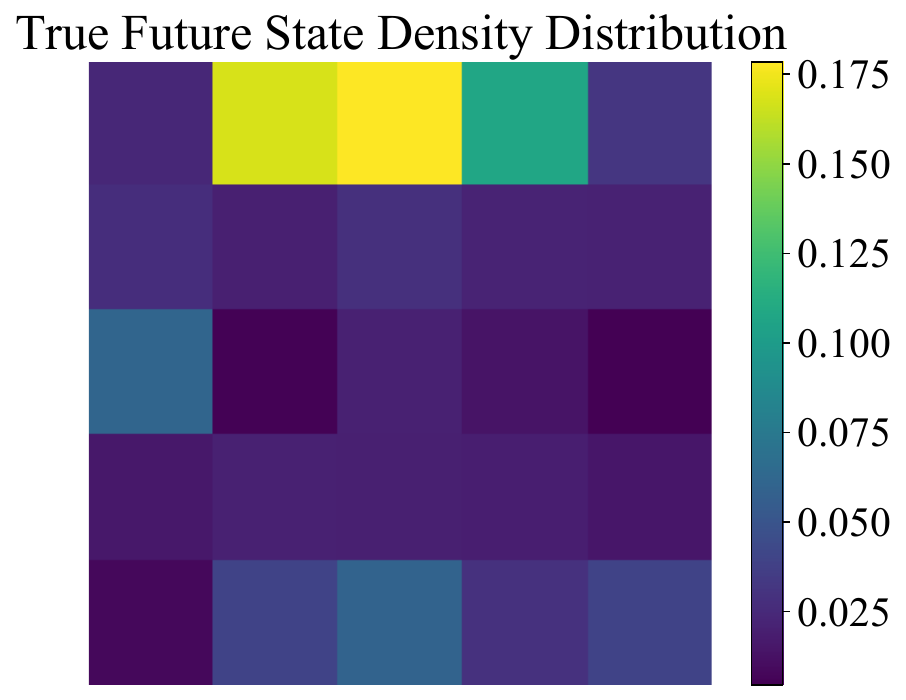}}
  \end{minipage}
  \begin{minipage}{0.32\linewidth}
		\centerline{\includegraphics[width=0.88\textwidth]{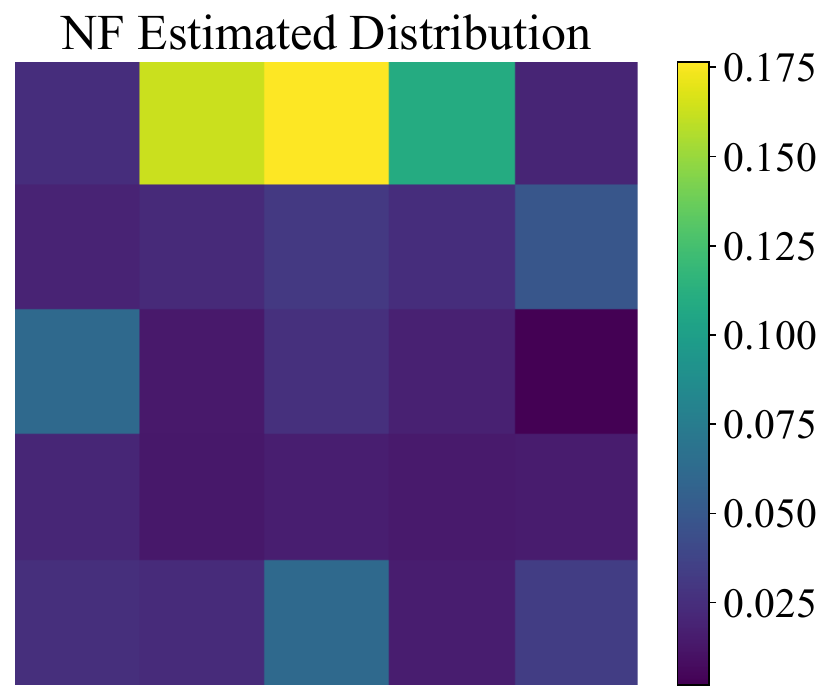}}
  \end{minipage}
  \caption{{\footnotesize \textbf{Experiments on the effectiveness of density estimation using Normalizing Flows.} \textbf{Left:} We evaluate CVAE, C-learning, CRL and Normalizing Flows for predicting the future state distribution in the on-policy setting. As anticipated, Normalizing Flows demonstrated the lowest estimation error among all methods evaluated. Conversely, CVAE exhibited the poorest estimation accuracy. In our empirical implementation, we observed that CVAE incurs significantly higher computational complexity due to its requirements for pre-training and importance sampling-based inference procedures \citep{wu2022supported}. \textbf{Middle:} and \textbf{Right:} The visual comparison. For a given state, action, and future goal in the GridWorld trajectory data, we visualize the comparison between the actual future state density (goal-reaching probability) and the estimates provided by the Normalizing Flows. The results indicate a minimal difference, further validating the effectiveness of the Normalizing Flows in estimating the future state density (goal-reaching probability).}}
  \label{fig:gridworld}
\end{figure*}
\paragraph{Analytic Future State Distribution} Then, as described in \citet{eysenbach2020c}, we can compute the true discounted future state distribution by first constructing the following two metrics:
\begin{align*}
    T \in \mathbbm{R}^{25 \times 25}: \quad &T[s, s'] = \sum_a \mathbbm{1}(f(s, a) = s') \pi(a \mid s) \\
    T_0 \in \mathbbm{R}^{25 \times 4 \times 25}: \quad &T[s, a, s'] = \mathbbm{1}(f(s, a) = s'),
\end{align*}
where $f(s, a)$ denotes the deterministic transition function.
The future discounted state distribution is then given by:
\begin{align*}
    P &= (1 - \gamma) \left[T_0 + \gamma T_0 T + \gamma^2 T_0 T^2 + \gamma^3 T_0 T^3 + \cdots \right] \\
    &= (1 - \gamma) T_0 \left[ I + \gamma T + \gamma^2 T^2 + \gamma^3 T^3 + \cdots \right] \\
    &= (1 - \gamma) T_0 \left(I - \gamma T \right)^{-1}
\end{align*}
The tensor-matrix product $T_0 T$ is equivalent to \texttt{einsum}(`ijk,kh $\rightarrow$ ijh', $T_0$, $T$).
We use the forward KL divergence for estimating the error in our estimate, $D_{\mathrm{KL}}(P||Q)$, where $Q$ is the tensor of predictions:
\begin{equation*}
    Q \in \mathbbm{R}^{25 \times 4 \times 25}: \quad Q[s, a, g] = q(g \mid s, a).
\end{equation*}

Following the configuration outlined in \citet{eysenbach2020c}, we compare the accuracy of the future discounted state distribution under against C-Learning and $Q$-learning:
\paragraph{On-policy Setting} \cref{fig:gridworld} presents the results of our evaluation comparing CVAE, C-learning, CRL and Normalizing Flows on the above modified "continuous GridWorld" environment under the on-policy setting.
In this scenario, CVAE demonstrates higher error compared to C-learning, while Normalizing Flows achieves the best performance. This highlights the accuracy of Normalizing Flows in estimating the discounted state occupancy measure. This experiment aims to answer whether Normalizing Flows solve the future state density estimation problem.
\clearpage
\subsection{Training Curves on Goal-conditioned Datasets from \citet{ghugare2024closing}}
\begin{figure*}[h]
    \centering
    \begin{minipage}{\linewidth}
		\vspace{3pt}
		\centerline{\includegraphics[width=0.5\textwidth]{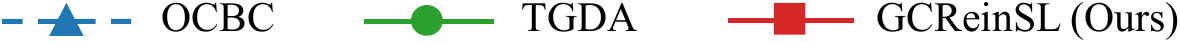}}
    \end{minipage}
    \begin{minipage}{0.32\linewidth}
		\vspace{3pt}
		\centerline{\includegraphics[width=\textwidth]{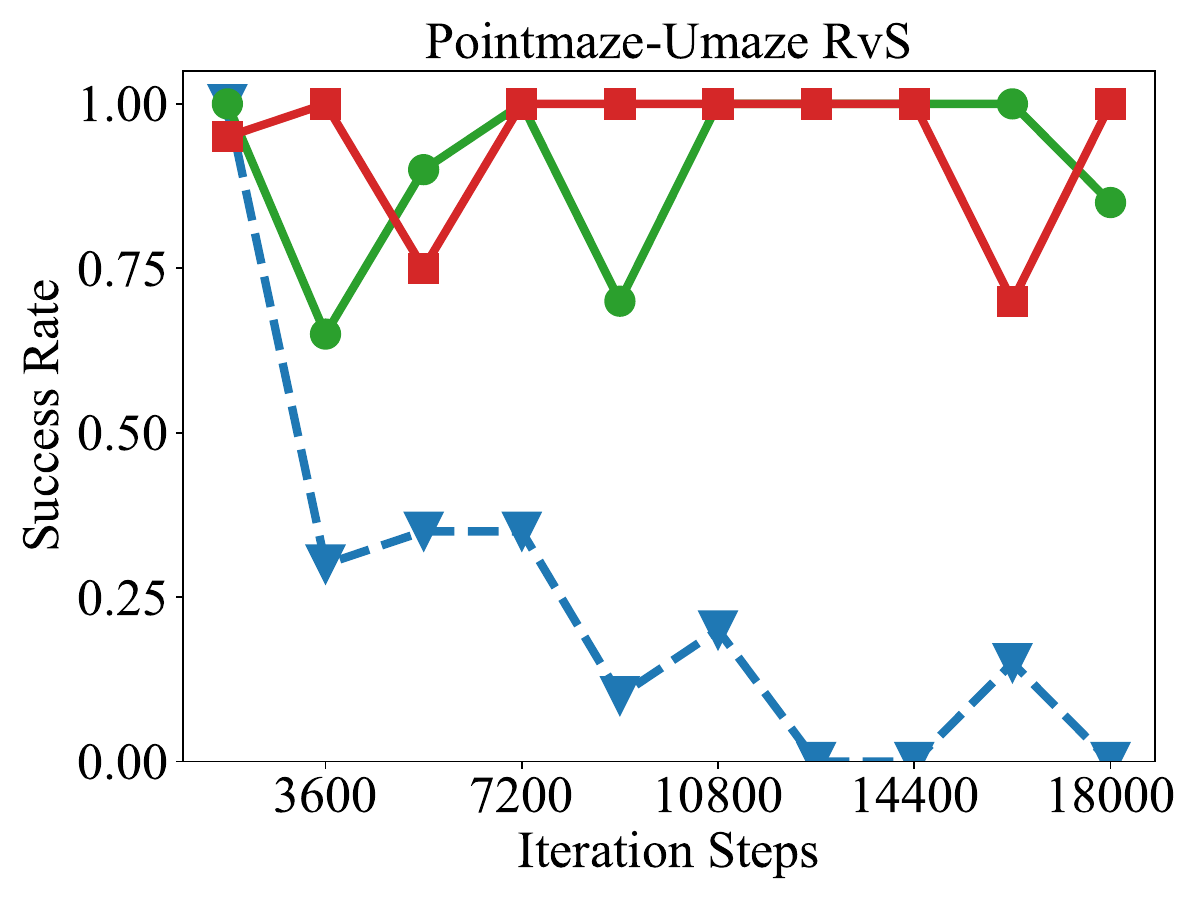}}
	\end{minipage}
	\begin{minipage}{0.32\linewidth}
		\vspace{3pt}
		\centerline{\includegraphics[width=0.98\textwidth]{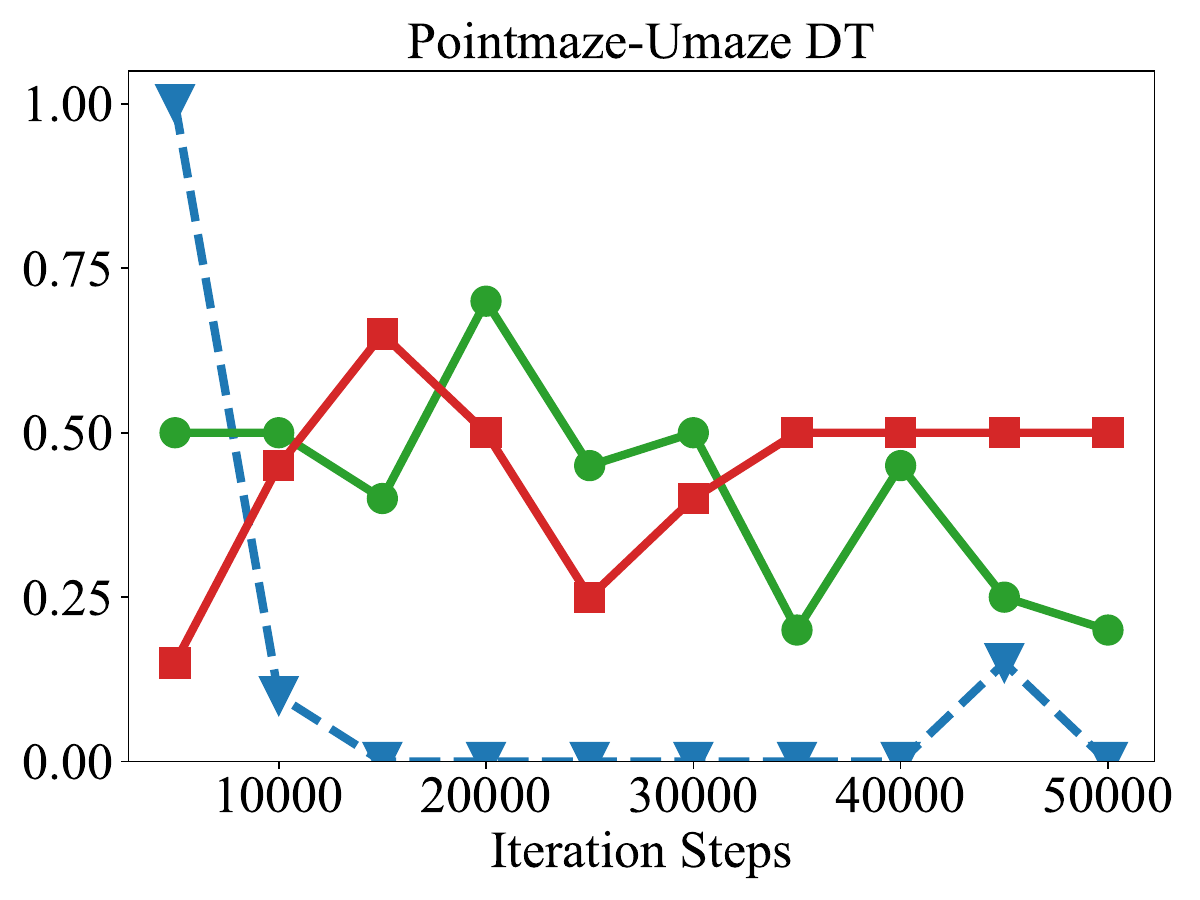}}
	\end{minipage}
    \begin{minipage}{0.32\linewidth}
		\vspace{3pt}
		\centerline{\includegraphics[width=0.98\textwidth]{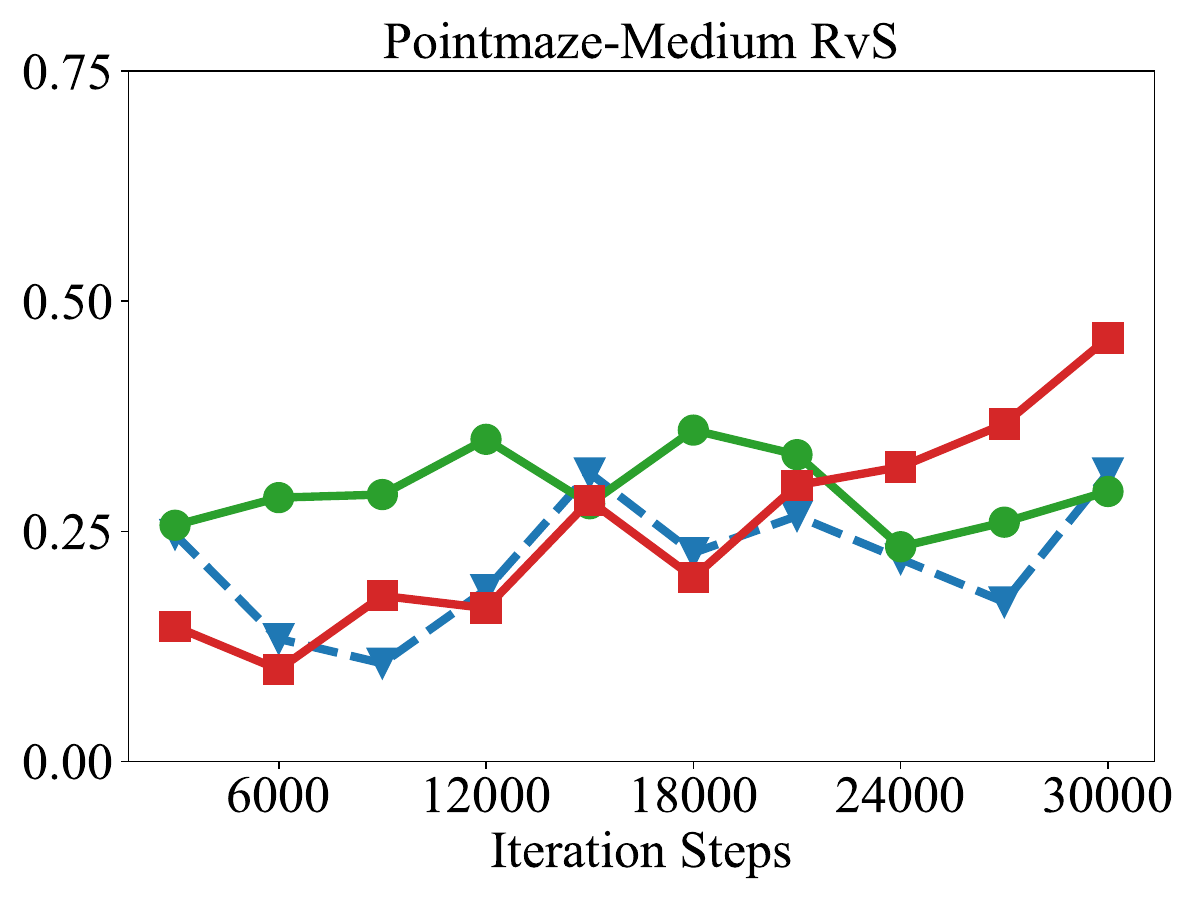}}
	\end{minipage}
    \begin{minipage}{0.32\linewidth}
		\vspace{3pt}
		\centerline{\includegraphics[width=\textwidth]{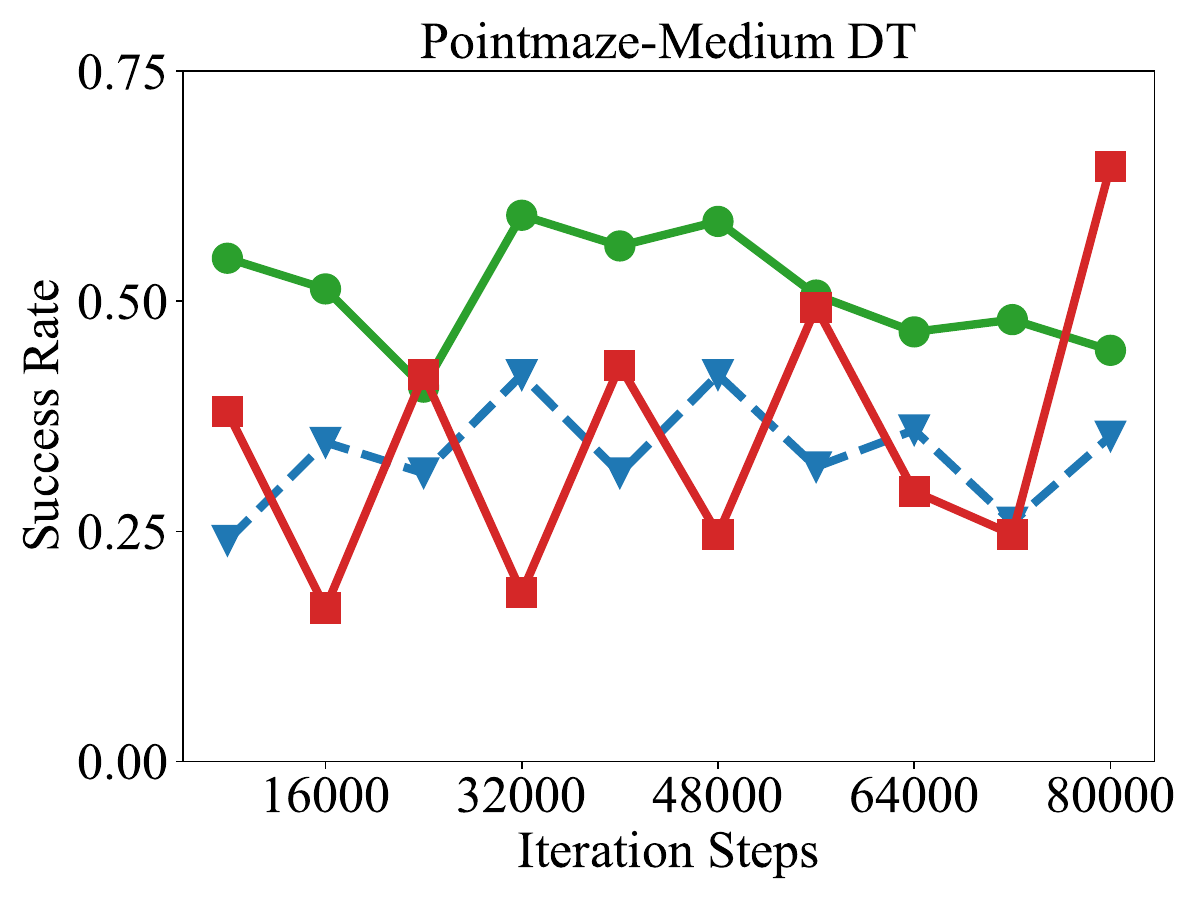}}
	\end{minipage}
	\begin{minipage}{0.32\linewidth}
		\vspace{3pt}
		\centerline{\includegraphics[width=0.98\textwidth]{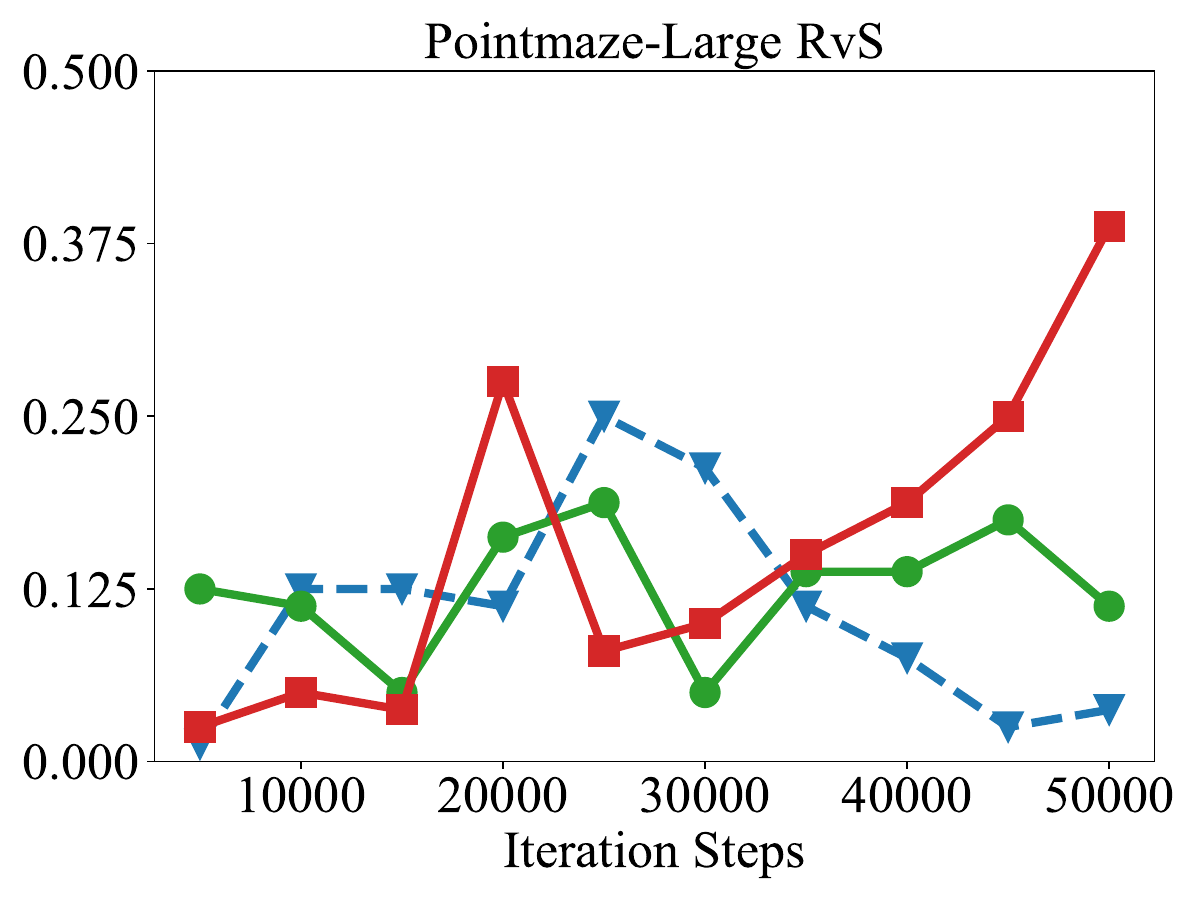}}
	\end{minipage}
    \begin{minipage}{0.32\linewidth}
		\vspace{3pt}
		\centerline{\includegraphics[width=0.98\textwidth]{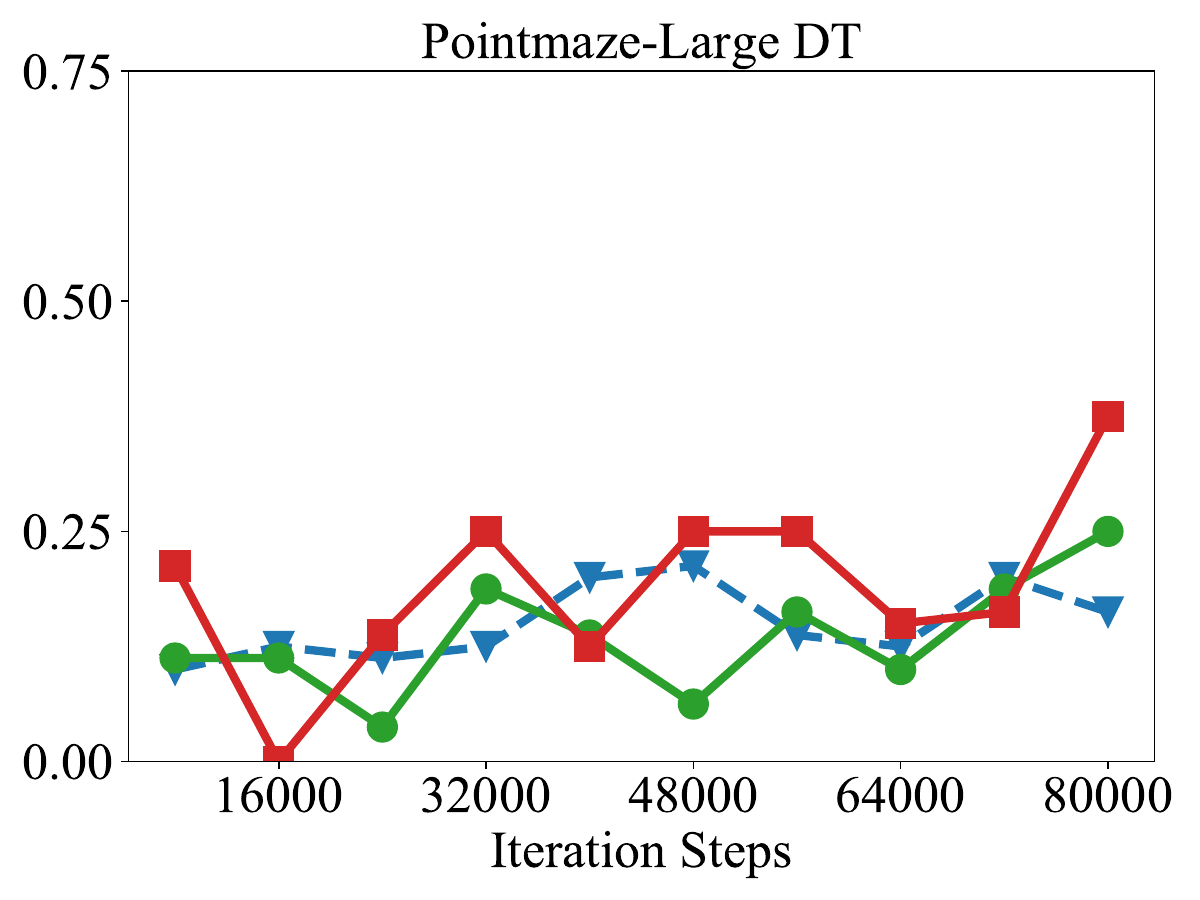}}
	\end{minipage}
    \begin{minipage}{0.32\linewidth}
		\vspace{3pt}
		\centerline{\includegraphics[width=\textwidth]{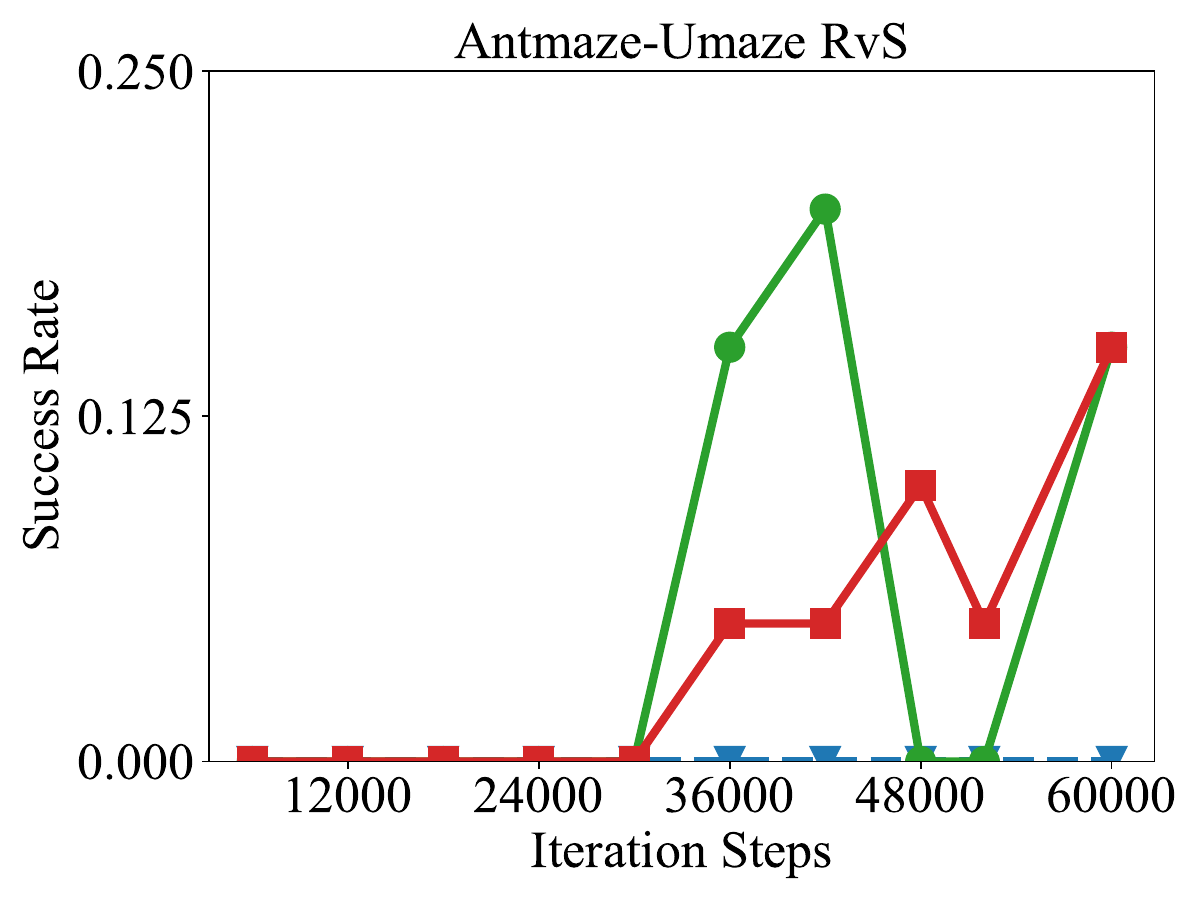}}
	\end{minipage}
	\begin{minipage}{0.32\linewidth}
		\vspace{3pt}
		\centerline{\includegraphics[width=0.98\textwidth]{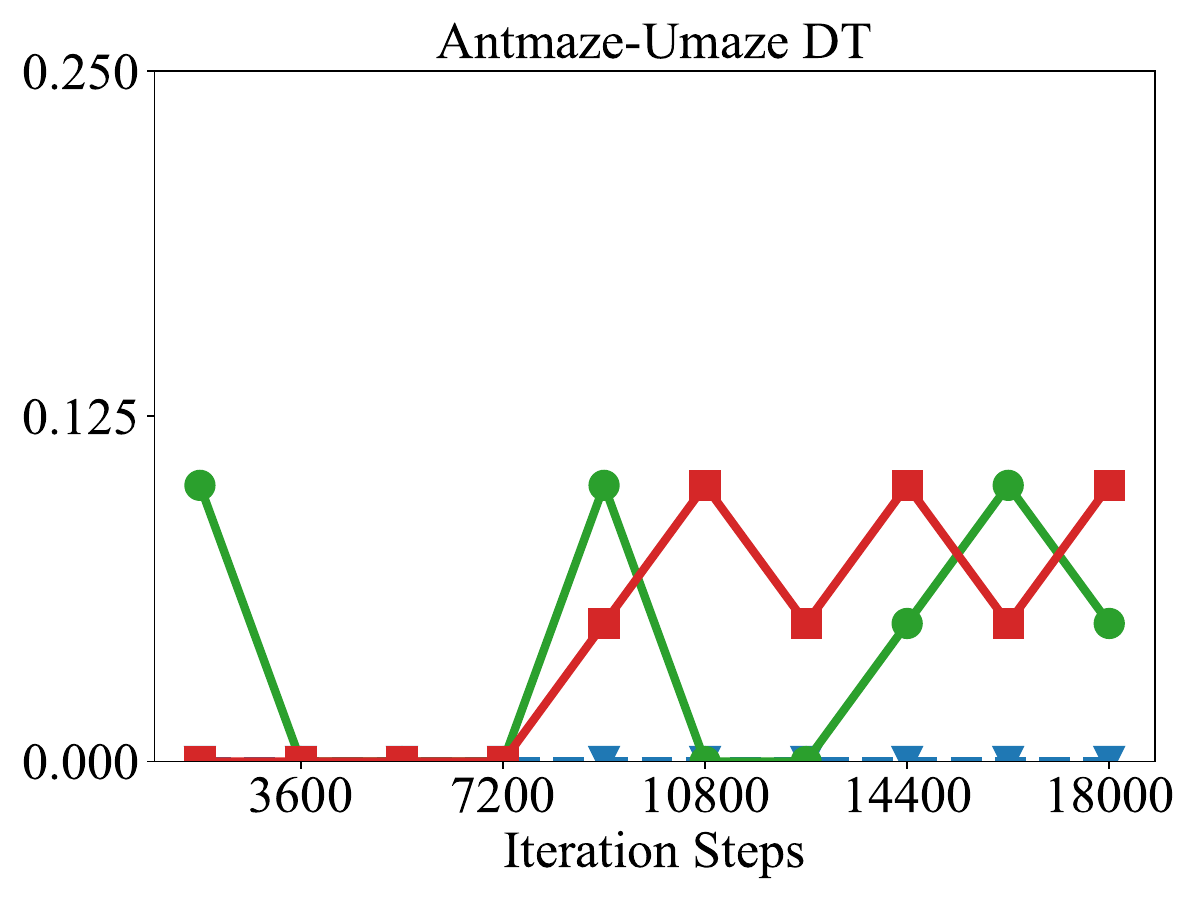}}
	\end{minipage}
    \begin{minipage}{0.32\linewidth}
		\vspace{3pt}
		\centerline{\includegraphics[width=0.98\textwidth]{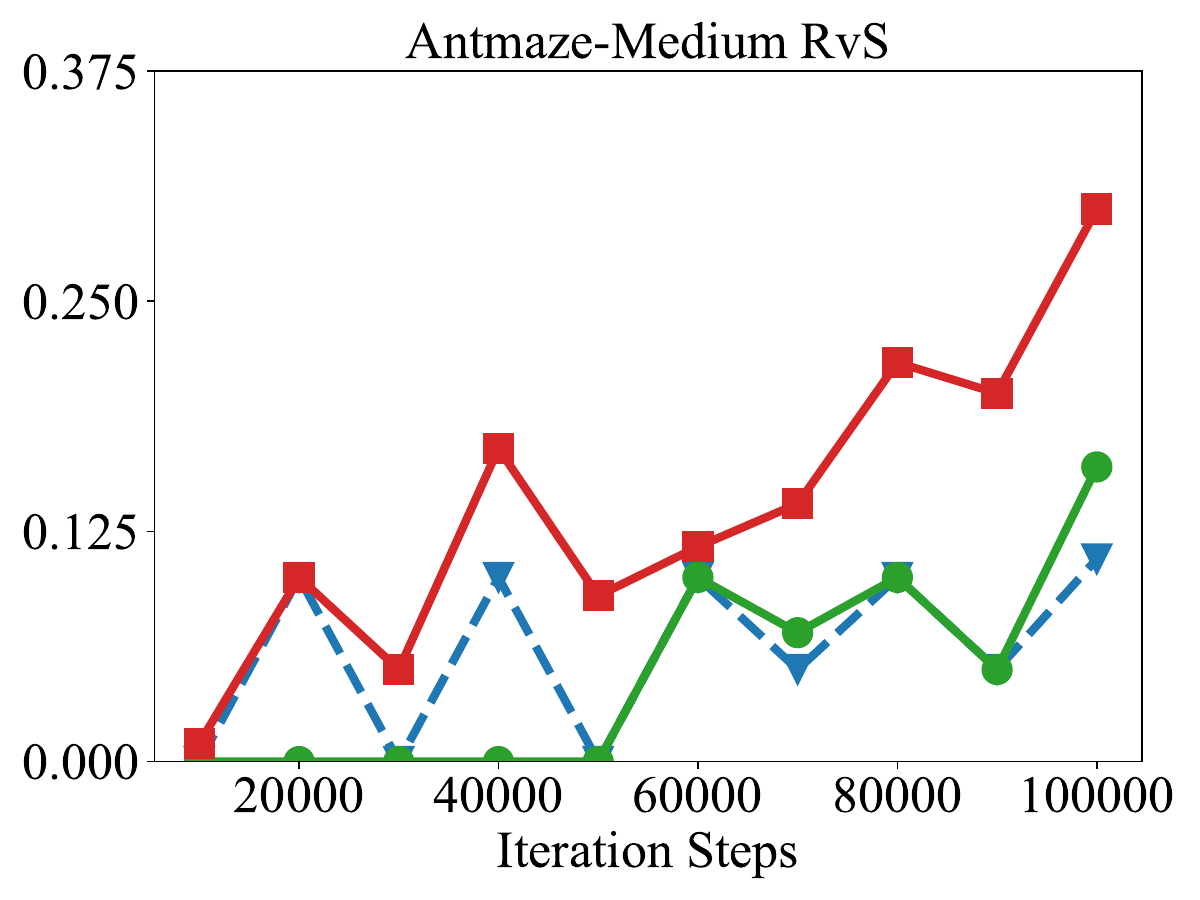}}
	\end{minipage}
    \caption{
    Training curves of OCBC and related goal data augmentation methods on \citet{ghugare2024closing} datasets. Although our \textbf{GC\textit{Rein}SL} method exhibits some instability on certain datasets, on average, \textbf{GC\textit{Rein}SL} tends to improve and achieves promising results with extended training. A potential direction for future research is to develop a more robust \textbf{GC\textit{Rein}SL} method that requires less hyperparameter tuning.
    }
    \label{fig:training curves results}
\end{figure*}

The training curves for nine datasets from \citet{ghugare2024closing} are shown in 
\cref{fig:training curves results}. The training process for \texttt{Pointmaze-Umaze} exhibits relatively stable behavior. However, the training on \texttt{Pointmaze-Medium} and \texttt{Pointmaze-Large} is characterized by high variance and significant fluctuations. Similarly, the \texttt{Antmaze-Umaze} dataset exhibits some degree of instability. Additionally, the performance on this dataset is notably poor. In contrast, performance on the \texttt{Antmaze-Medium} dataset shows a stable improvement, with the trends for \textbf{GC\textit{Rein}SL} for DT and \textbf{GC\textit{Rein}SL} for RvS aligning closely. On the \texttt{Antmaze-Large} dataset, the majority of average success rates are near zero.

\clearpage

\section{Limitations} \label{dis:limatation}
The proposed framework has several limitations. 
First, the performance of \textbf{GC\textit{Rein}SL} is highly dependent on the accuracy of the estimated discounted state occupancy distribution. For instance, when an estimator such as a CVAE is employed, the performance may deteriorate significantly.

Secondly, while SL methods, such as sequence modeling, are straightforward and efficient, their actual performance still falls short compared to classical RL approaches. Moving forward, it is essential to develop more advanced SL methods that not only surpass the performance of traditional RL techniques but also fully exploit the advantages inherent in SL. For example, by integrating our $Q$-conditioned maximization with Decision Mamba \citep{lv2024decision,ota2024decision,huang2024decision,cao2024mamba,zhuang2025revisiting}.

\section{Societal Impact} \label{dis:impact}
This paper presents research aimed at advancing the field of RL. This research is centered on enhancing the stitching capability in the field of offline
reinforcement learning: OCBC methods. By overcoming their limitations, it contributes to
the advancement of offline reinforcement learning. As foundational research in machine learning,
this study does not lead to negative societal outcomes.

\end{document}